\documentclass[journal]{IEEEtran}

\usepackage[colorlinks,urlcolor=blue,linkcolor=blue,citecolor=blue]{hyperref}

\usepackage{amsmath,amssymb,amsfonts,physics}
\usepackage{color,array}

\usepackage{subfigure}
\usepackage{graphicx}
\usepackage[normalem]{ulem}

\newtheorem{remark}{Remark}
\usepackage{subfigure}

\newcommand{\scr}[1]{\ensuremath{\mathcal{#1}}}\newcommand{\comment}[1]{}

\begin{document}

\title{Encoder-Decoder RNNs for Bus Arrival Time Prediction}

\author{Nancy Bhutani$^{\S}$,  Soumen Pachal$^{\S}$, and Avinash Achar*$^{\S}$ 
\thanks{* Avinash Achar is the corresponding author.}
\thanks{$^{\S}$ All authors contributed equally.}
\thanks{ Nancy Bhutani, Soumen~Pachal, Avinash~Achar    are with TCS Research, Chennai, INDIA.  E-mail: \textit{\{nancy.9,s.pachal,achar.avinash\}@tcs.com.} }

\thanks{The manuscript was first submitted in Aug 2023 for review.}
\thanks{This paragraph will include the Associate Editor who handled your paper.}}

\markboth{IEEE }
{N Bhutani \MakeLowercase{\textit{et al.}}}

\maketitle

\begin{abstract}
Arrival/Travel times for public transit exhibit variability due to factors like seasonality, traffic signals, travel demand
fluctuation etc. The developing world in particular is plagued by additional
factors like lack of lane discipline, excess vehicles, diverse modes of transport, unreliable schedules etc. This renders the bus arrival time
	prediction (BATP) to be a challenging problem especially in the developing world. A data-driven model based on a novel variant of Encoder-Decoder (OR Seq2Seq) recurrent neural networks
(RNNs) is proposed for BATP (in real-time). The model
intelligently incorporates  spatio-temporal (ST) correlations in a unique
	(non-linear) fashion distinct from existing approaches. Existing Encoder-Decoder (ED) approaches for BATP blindly map time to the sequential aspect of ED, while ignoring crucial data characteristics and
        making some restrictive model assumptions. Our approach in contrast is not straightforward and effectively tackles these issues.
We exploit the geometry of the dynamic real-time BATP problem to enable a  novel fit with the
ED structure, distinct from existing ED approaches.
Further motivated
from accurately modelling past congestion influences from downstream sections, we additionally propose  a bidirectional layer at the decoder (something unexplored in other time-series based ED application contexts). The effectiveness of the proposed architecture
is demonstrated on
        real field data collected from challenging traffic conditions, while bench-marking against state-of-art baselines. The proposed architecture is not limited to transportation, but can also be employed for multi-step time-series forecasting (sales/demand forecasting under exogenous inputs like price).  

\end{abstract}

\comment{
\begin{IEEEImpStatement}
BATP continues to be an important and challenging problem  in the developing
world.  Our novel ANN architecture performed better than a range of existing state-of-art approaches with improvements of up-to 13\%.  Our methodology can be a potential component of Advanced
Traveller Information Systems (ATIS) which provides  variety of real-time information for
commuters. Enhancing  public transit standard can motivate commuters to avoid hitting 
the road  individually. This can reduce traffic volumes, in
turn mitigating pollution. Further, accurate BATP estimates can aid commuters to (i) plan their  bus-
stop arrival,   (ii) decide whether to choose a bus at all for commute based on
predictions.  
Our  novel ANN architecture for BATP, which also incorporates influences of past congestions from downstream sections goes beyond the transportation application. It  can be employed for multi-step sales prediction in domains like retail, CpG (Consume
packaged goods) etc. where it can model anti-causal influences of future promo prices on current sales.
\end{IEEEImpStatement}
}

\section{Introduction}

Public transit system  is a crucial component to administer the overall transport system in urban cities across the world. Having a quality system would make it attractive for commuters and can in-turn mitigate mounting traffic volumes and congestion levels, which is a universal problem across the urban world. A quality system would entail sticking to well-designed schedules to the extent feasible while providing quality  predictions in real-time. Such accurate estimates can help commuters better plan their bus-stop arrival  
while avoid unnecessary wait times. Quality Bus travel time predictions can also assist commuters  decide between taking a bus or some other alternate mode of transport. Quality BATP estimates can also benefit travel administrators take corrective action when bus schedules are violated. 

\comment{
 Constantly mounting traffic congestion levels is a universal problem across geographies on account of steadily increasing urban population and
associated traffic volumes. Enhancing the public transit standard is one promising  approach to mitigate this ubiquitous problem. 
This not only reduces traffic volumes and
congestion in turn, but would also curtail pollution.
 Public transit needs to be reliable to remain attractive among commuters.    Such accurate estimates can aid commuters plan their arrival to bus-stops and minimize waiting times.   Quality Bus travel
time predictions can also assist commuters  decide whether to take a bus or some alternate mode of transport. The  bus arrival time prediction (BATP) can potentially 
also aid transport administrators take real-time corrective measures when the bus is off schedule. Hence providing quality BATPs is  crucial for reliable mass
transit.
}

BATP literature is more than one and a half decades old.  It continues to be a challenging
research problem in the 
developing world, in particular.
 Factors contributing to this include  (1) absence of lane discipline (2) in-homogeneity of
traffic (i.e.  transport modes can include bicycles, two wheelers, four wheelers, heavy vehicles like trucks, buses and so on) with dedicated lanes absent for specific modes 
of transport.
We refer to this as
{\em mixed traffic} conditions. Another issue is the lack of reliability of bus schedules \cite{ITSC20}, especially in the context of India where most cities experience mixed traffic conditions. 
The bus schedules even if available tend to get outdated due to constant changes in traffic conditions and infrastructure. This makes timetables extremely unreliable, leading to ad-hoc waiting times for passengers. Hence providing accurate arrival time predictions become even more important in such cases.
The real data considered in this paper is from  a bus route in Delhi \cite{ITSC20}, the capital of India which experiences mixed-traffic conditions.    
Any google-map based bus arrival time query  (in most cities in India including Delhi) has till recently mostly returned a constant prediction  independent of the 
date or time of
query. These constant estimates  seem to be based on some pre-fixed schedules which  are unfollowable given the complex  traffic conditions as explained above. 
Further, ETA (Expected time of arrival) solutions are still being continuously improved upon by Google at a network level across the world for different modes of urban transport \cite{GoogleETA21,KDD20}.
On account of the above factors, 
BATP continues to be a challenging research problem \cite{ranji19} especially under mixed traffic conditions \cite{Achar19a,Paliwal19}.

Over the years, there have been  diverse approaches proposed for BATP. Data-driven approaches have been a dominant class of methods for BATP. 

In most of these approaches, an entire route is segmented into smaller sections(or segments) either uniformly \cite{lelitha:09} OR into non-uniform segments connecting 
consecutive bus-stops. 
Depending on the method and the installed sensing infrastructure, the data input  can be  entities like speed, density,  flow,  travel time etc.  
In this work, we consider   
scenarios where input data  comes only from  travel times experienced across these segments/sections. 
AVL (automatic vehicle location)  data captured by GPS sensing can {\em easily} provide  such travel times. 

\textbf{Gaps and Contributions:} Over the years, researchers have explored a wide spectrum of methods under the broad umbrella of data-driven approaches.  These include ARIMA models \cite{Jairam18}, linear statistical models like Kalman filters
\cite{Achar19a,Dhivya20},  support vector machines \cite{bin:07,reddy:16,Achar19b}, feed-forward ANNs \cite{fan:15,KDD20}, 
recurrent neural networks \cite{Petersen19,Ran19}, CNNs \cite{Paliwal19}, temporal difference learning\cite{Vignesh20} and so on. Most of the existing methods suffer from a range of 
issues 
like (i)insufficient utilization of historical  
data for model calibration \cite{snig:15,partc:17,lelitha:09} OR  (ii)ignoring   spatial correlations \cite{Dhivya20,vivek:17,reddy:16}  
(iii) not exploiting temporal correlations(\cite{fan:15,Duan16}) (iv)
not exploiting real-time information enough(\cite{yang:16}), (v)segment the time-axis into uniform bins, which can lead to inaccurate 
predictions\cite{Petersen19,Ran19}.  There has been some recent work on exploiting 
spatio-temporal correlations \cite{Achar19a,Achar19b,Paliwal19,Petersen19,Ran19} as well in this direction. 

From an RNN literature perspective, there's been recent work where people have explored ED (also known as Seq2Seq) architectures for real-valued data
\cite{Wen17,Yagmur17} (time-series (TS) in particular). 
A natural way to employ Seq2Seq for BATP would be to segement the time-axis into uniform bins  \cite{Petersen19,Ran19} and learn a sequential model in time.
This approach ignores the continuous nature of data in the temporal dimension and  makes an unrealistic assumption that section travel times are constant across time bins (see Remark $2$).
Our approach intelligently addresses this drawback while also respecting the temporal continuity and space-time asymmetry, in BATP data.
Further, proposed approach is not evident due to difference in structure of available data between BATP and traditional time-series.

\comment{
However, employing a Seq2Seq (ED) architecture for BATP as prescribed in these time-series approaches is not immediately evident due to 
difference in 
structure of 
available data between BATP and traditional time-series. In particular, GPS based data has a unique spatio-temporal aspect with information scattered 
across trips throughout the day.  
}

The current work proposes a novel ED architecture (different from classical machine translation architecture OR existing 
ED approaches for time-series)  
which is also distinct from all existing BATP approaches (in particular from ED based BATP approaches also \cite{Petersen19,Ran19}). It exploits current real-time 
spatio-temporal correlations and historical seasonal  correlations   for nonlinear modelling and  prediction. Specifically our contributions are as follows.
\begin{itemize}
	\item {\em We intelligently recast the real-time BATP problem to a novel and unique  variant of Encoder-Decoder architecture with real-time spatio-temporal inputs and historical 
		seasonal inputs carefully placed in the architecture. The key is in recognizing that BATP  inherently involves sequential training data with variable length input-output pairs, which ED framework can handle. The proposed ED model's sequential aspect  is mapped to the spatial  dimension of BATP, while the temporal aspects of BATP are captured by feeding the associated inputs as decoder inputs in a space synchronized fashion.} 
		
				 	
	\item Travel times from the just traversed sections of current bus (constitute the spatial correlations) are fed as inputs to the
		encoder. While  real-time information coming from closest previous bus's travel times across subsequent sections  (constituting 
		temporal correlations) are fed in a synchronized sequential fashion into the decoder as additional inputs.  {\em Note that these synchronised inputs at the decoder are absent in the classic ED application for machine translation \cite{Cho14,Sutskever14}.  The section travel times (of current bus) across subsequent sections are the prediction targets which are neatly mapped to the decoder outputs sequentially. } Weekly seasonal correlations are also incorporated via additional inputs from the closest trip of the previous week.
	\item {\em We propose a bidirectional layer at the decoder as this can now capture (for a given section) the possible (upstream) influence of past 
		congestions (in time) from subsequent sections, propagating backward in space (along the bus route).  This novel feature of our ED variant is unexplored
		in both traditional ED and other time-series based variants of ED to the best of our knowledge. }
	\item  Effectiveness of the  proposed approach is illustrated on actual field data collected from a route in Delhi, where  mixed traffic condition 
		are very common. The results  
		clearly demonstrate superior performance of our approach  (across sub-routes of diverse lengths)
		in comparison to $4$ recent state-of-art baselines. 
\end{itemize}

The rest of the paper is organized as follows. Sec.~\ref{sec:RelatedWork} describes the related work in detail both from the perspective of  (i) BATP literature and (ii) ED based RNN approaches. Sec.~\ref{sec:Methodology} explains the technical contribution in detail. In particular it describes the proposed architecture and technically motivates  how the architecture can be derived to solve BATP. Sec.~\ref{sec:Results} demonstrates the effectiveness of the proposed architecture on one route from Delhi by bench-marking against four carefully chosen state-of-art baselines. We provide (i) a brief discussion as to how our proposed architecture can be also be used in other applications  and (ii) concluding remarks in Sec.~\ref{sec:Conclusion}

\section{Literature Review and Related Work}
\label{sec:RelatedWork}
The BATP literature has not only seen diversity in the range of techniques used, but also in the kind of data input employed for prediction. 
A range of data inputs like  speed, travel times, weather, flow information,   crowd-sourced
data\cite{zhou:12}, scheduled time tables \cite{lin:99}  and so on have been considered  for BATP.
One can broadly categorize the range of approaches into two classes: 
(i) traffic-theory based  (ii) data-driven. Given the proposed approach is data-driven, we stick to reviewing related data-driven approaches. 
While Sec.~\ref{sec:BATP} describes  data-driven approaches for BATP, Sec.~\ref{sec:RNNbased} discusses related ED based RNN approaches. Finally, Sec.~\ref{sec:Perspective} places the proposed architecture in perspective of all related work.

\vspace{-0.05in}
\subsection{Data-driven methods for BATP}
\label{sec:BATP}
 Unlike the traffic-theory based approaches which model the  physics of the traffic,  data-driven methods employ a 
coarse model (based on measurable entities) that is sufficient for predictive  purposes. 
 Most  approaches learn to estimate necessary parameters of a suitable predictive model from   past historical 
data, which is further  employed for real-time BATP.  There are a few approaches  which employ a
data-based model but do not perform  full-fledged learning based on historical data.

{\bf Without Learning:} 
 One of the  earliest approaches without learning  was proposed in \cite{shalaby:03} using a Kalman filter. As inputs, it used   previous bus travel times and travel times from previous day (same time).  It has an arbitrary choice of  parameter in its state space model while only capturing temporal dependencies.  
The  subsequent approaches consider a linear state-space model involving travel times and   {\em calibrate} (or fix) the data-based model parameters in real time. They choose their 
parameters  either based on (a) 
data from previous bus \cite{snig:15,lelitha:09} or (b)an appropriate optimal travel-time data vector from the historical data-base \cite{kumar:12}.

{\bf Explicit Learning:} As mentioned in the introduction, there are a  variety of  approaches learning from historical data.  For instance,  
support vector regression \cite{reddy:16} and  feed-forward 
ANNs \cite{vivek:17} were employed  
to capture temporal correlations via multiple  previous bus travel times.  
Employing  link length (a static input) and rate of road usage and speed (dynamic inputs),  \cite{yang:16} proposes an SVR based prediction. 
However  current bus position   OR previous bus inputs are not considered there. 
 A speed based prediction scheme is proposed in  \cite{sun:07} which uses a weighted average of current bus speed and historically averaged section speed as inputs. As previous method, it ignores
information from previous bus.  
 A dynamic SVR based prediction scheme is proposed in \cite{bin:07} which exploits spatio-temporal (ST) correlations in a minimal manner.  In particular, it considers current bus 
 travel time at the previous section and previous bus travel time at the current section. 


A single feed-forward ANN model is built to predict travel times between any two bus stops on the route in \cite{fan:15}. On account of this, target travel time variable's dynamic range would be very large and can lead to poor predictions for very short and very long routes.
An approach using (non-stationary) linear
statistical models which captures ST correlations  was proposed in \cite{Achar19a}. It uses a linear kalman filter for prediction.  Linear models here are used to 
capture spatial correlations. The temporal correlations come from the (currently plying) previous bus section travel time.  Another approach using linear statistical models  and exploiting real-time temporal correlations (from previous buses) was proposed in \cite{Dhivya20}. A nonlinear generalization of \cite{Achar19a} using support vector function 
approximators capturing ST correlations was proposed in \cite{Achar19b}.  Recently, a CNN  approach capturing ST correlations 
was proposed
in \cite{Paliwal19}.  It uses masked-CNNs to parameterize the predictive distribution, while a quantized travel-time is used as CNN outputs.

\cite{Petersen19} proposes a novel approach by combining CNNs and RNNs in an interesting fashion. In particular spatial correlations from the 
adjacent sections of the 1-D route are captured by the convolutional layer,
while the recurrent structure captures the temporal correlations. It employs a convolutional-RNN based 
ED architecture 
to make multi-step predictions in time. \cite{Ran19} considers an attention-based extension of  \cite{Petersen19}. \cite{Wu20} employs a simplified RNN with attention
but no state feedback (even though weight sharing is present across time-steps). It only captures single time-step predictions. A common feature of all these RNN approaches is that the time axis is 
uniformly  partitioned into  time bins of a fixed width.

A recent computationally interesting approach where BATP is recast as a value function estimation problem under a suitably constructed Markov reward process is proposed in \cite{Vignesh20}. This enables 
exploring a  family of value-function predictors using temporal-difference (TD) learning.
 
\subsection{\bf Related ED based RNN approaches}
\label{sec:RNNbased}
The ED architecture was first successfully proposed for language translation applications\cite{Cho14,Sutskever14}. The proposed architecture was relatively simple with the
context from the last time-step of the encoder fed as initial state and explicit input for each time-step of the decoder. Over the years, machine
translation literature has seen intelligent improvements over this base structure by employing attention layer, bidirectional layer etc. in
the encoder.  Further, the ED framework has been successfully applied in many other tasks like speech recognition\cite{Liang15}, image captioning etc.  

Given the variable length Seq2Seq mapping ability, the ED framework naturally can be utilized for multi-step (target) time-series prediction where
the raw data is real-valued and 
target vector length  can be independent of the input vector. An attention based ED approach (with a bidirectional layer in the encoder) for 
multi-step TS prediction was proposed in
\cite{Yagmur17} which could potentially capture seasonal correlations as well. However, this architecture doesn't consider exogenous inputs. An
approach to incorporate exogenous inputs into predictive model was proposed in \cite{Wen17}, where the exogenous inputs in the forecast horizon are fed
in a synchronized fashion at the decoder steps. Our approach is close to the above TS approaches. 

\comment{     
PLEASE NOTE
Para on existing ED approaches for real-valued data, in particular time-series applications. This way the related work comes from both BATP and RNN for time-series. And then explain (in the next subsection) how the proposed approach is
different from these TS based approaches. Trim the approaches from BATP literature. KF-enhanced can certainly go.   Also include the gist of
these ED approaches even under gaps and contributions and mention how our ED architecture is different from existing TS ED approaches AND
also from existing BATP approaches. 
}

\subsection{\bf Proposed approach in perspective of related approaches}
\label{sec:Perspective}
From the prior discussion, one can summarize that  many existing approaches either fail to exploit historical data sufficiently OR fail to capture  
spatial or temporal correlations.
The rest of the approaches do exploit spatio-temporal correlations in different ways \cite{bin:07,Achar19a,Achar19b,Paliwal19,Petersen19,Ran19}, but suffer their own 
drawbacks. For instance, \cite{bin:07} while exploits the previous bus travel time at the current section   (temporal correlation), completely ignores  
when (time of day) the  traversal happened. The spatial correlation here comes from  current bus travel time of only one  previous section. 
\cite{Achar19a}  (denoted as LNKF in our experiments) addresses the issues of \cite{bin:07} as follows. To better capture spatial correlations, it considers current
bus travel time measurements from multiple previous sections. The temporal correlations here also take into account the previous bus's proximity    
by assuming a   
functional (parameterized) form 
dependent on current section travel time and start time difference. It adopts a predominantly linear modelling approach culminating in a
Linear Kalman filter for prediction. As explained earlier, a support-vector based  nonlinear generalization of \cite{Achar19a} is considered in
\cite{Achar19b} (referred to as SVKF in our experiments). It learns the potentially non-linear spatial and temporal correlations at a single-step level and then employs an extended kalman
filter for spatial multi-step prediction. Compared to our non-linear ED (Seq2Seq) approach here, \cite{Achar19a} considers mainly a linear modelling. While
\cite{Achar19b} adopts a non-linear modelling, the model training happens with single-step targets in both \cite{Achar19a,Achar19b}.  {\em However, 
both these KF approaches adopt a
recursive sequential multi-step prediction which can be prone to error accumulation. On the other hand, our ED approach circumvents this issue of 
both these KFs by training with vector targets where the predictions across all subsequent sections are padded together into one target-vector.}

CNN approach of \cite{Paliwal19} models travel time targets as categorical values via a soft-max output layer. Hence it is sensitive to the quantization level. A coarse quantization can lead to high errors when the true target value is exactly between two
consecutive levels.  A fine quantization on the other hand leads to numerous outputs. This in turn would increase the  number of weights to be learnt and lead to a potentially  
imbalanced multi-class problem.   Our approach in contrast models targets as real-valued. 

All LSTM approaches \cite{Petersen19,Ran19,Wu20} bin the time-axis into fixed  width intervals.  
The method in \cite{Dhivya20} also bins the time-axis and builds temporal models at each section while only exploiting temporal correlations.
This strategy makes an inherent assumption that section travel times  
 are constant for a fixed width time interval ($15$ minute in particular). {\em However, this assumption can be pretty unrealistic and restrictive during peak hours (in particular) when the traffic conditions are more dynamic.       
Our approach on the other hand doesn't make any such restrictive assumptions and we model time as  continuous valued.}

While \cite{Petersen19} (referred to CLSTM in our experiments) also uses an ED architecture  for prediction, it employs sequential aspect of RNNs to model time (unlike our method). In contrast, 
our sequential RNN (ED) captures the spatial dimension 
of the problem.  Accordingly our decoder output models the subsequent section travel times   of the current bus. 
In contrast,  a $1$-D CNN is used in \cite{Petersen19} to capture spatial correlations.  To capture temporal correlations,  our method feeds the entry and travel times  at subsequent sections of 
the closest previous bus as decoder inputs in a space synchronised fashion. Further we also use a bidirectional layer at the decoder to capture possible
upstream propagating congestion influences. This makes our ED approach very different from CLSTM \cite{Petersen19}. {\em Overall, there is no symmetry in the spatial and temporal aspect of AVL data which makes our proposed ED variant very different from CLSTM.}

The TD approach of \cite{Vignesh20} (as explained earlier) where a value function estimation of a suitable Markov reward process (MRP) is carried out, is clearly  
distinct from our approach here. The travel-time across each section is modeled as the one-step reward. For each destination section (or bus stop), a MRP is constructed with all preceding sections (along with features like time of day etc.) encoded as states, while the destination section is the final state. {\em This approach while conceptually interesting performs poorly for short to mid-range sub-routes.} 

 BusTr \cite{KDD20} models the bus-stop dwell time and run time across a segment separately using a feed-forward ANN (FFN), where the FFN weights are shared across 
 road segments and bus stops. The inputs to the FFN are the (i)speed forecast based on current real-time traffic from google maps and (ii)features based on location embedding, 
 both of which are unused in our approach. The speed forecast used is that of cars, whose speed based travel times are linearly transformed to that of bus while the 
 linear weights are modelled as the output of the FFN whose weights are learnt during training. {\em Overall BusTr differs from our method in most aspects like the inputs 
 and targets used for prediction and the predictive model.}

Compared to the ED based TS prediction approaches \cite{Yagmur17,Wen17}, one novel feature is the use of a bidirectional layer at the decoder (unexplored to the 
best of our knowledge in TS approaches). This feature is strongly motivated from the BATP application as explained in detail in Sec.~\ref{sec:Bidir}. Also, 
in contrast to \cite{Yagmur17,Wen17}, we employ additional inputs at the decoder to factor real-time temporal and weekly seasonal
correlations, something absent in   \cite{Yagmur17,Wen17}.

\section{Methodology}
\label{sec:Methodology}
This section is organized as follows. Sec.~\ref{sec:DataPrelims} describes data preliminaries while Sec.~\ref{sec:STSCorrelations} describes the various potential correlations in the data that can influence BATP.  Sec.~\ref{sec:Prediction} states the dynamic real-time prediction problem and  describes the various inputs and notations. Sec.~\ref{sec:EDArch} 
describes how BATP can be intelligently recast into a novel ED framework with spatio-temporal and seasonal inputs carefully placed. 
\subsection{Data Preliminaries}
\label{sec:DataPrelims}
{\bf Data Input:}   The given bus route is segmented  uniformly into sections. 
Along the given route, travel times experienced across each of these sections for all concluded trips, constitute  our input training data. The observed section
travel times, include both (i)dwell time at the possible bus-stops in the section and (ii)running time across the section.  
As explained in the introduction, travel times were obtained from AVL data captured via high frequency GPS based sensing.

{\bf Uniform Segmenting:} We chose this for two reasons. (i)There is  prior work \cite{lelitha:09,partc:17} where researchers have adopted uniform segmenting and predicted 
 between any two bus stops, and
reported good performance. 
(ii){\em Ease of testing across multiple bus-routes:} Substantially extra book-keeping would be needed when Non-uniform segmenting is employed (where section ends correspond to bus-stop locations)  (iii) Ease of adapting section-level predictions to the bus-stop level as explained next.

{\em Our proposed approach can also be used as it is when the route is segmented in a non-uniform fashion  (where travel times between two successive bus-stops will be the section travel times).}  We strongly believe that the results 
in the non-uniform case will be similar in trend to the  uniform
case results reported here.

{\bf Adapt to bus-stop level:} Also, the predictions based on uniform segmenting which predict between any two segments can be easily adapted to predict between any two bus-stops. Specifically, given any two bus stops
$i$ and $j$, we could consider the segment start in which bus-stop $i$ lies  and perform our proposed ED based prediction till the end of the segment which contains bus-stop $j$. We
would now need to subtract the expected dwell time at bus-stop $j$ and the expected semi-segment travel times at the start and end segment of the multi-step prediction. The
semi-segment lengths will depend on the position of the bus-stop $i$ and $j$ in their
respective segments.

\comment{
The route The data used in this paper was collected from a bus route in Chennai, an Indian city. 
All the plying buses were GPS
enabled with their position information logged every few sec. The bus route was segmented into sections  each of 400 m length and time taken to 
cover each section was calculated by linear interpolation from the high frequency GPS data.  The $400$ m choice stems from the fact that the average distance between bus-stops on
the chosen route being slightly above $400$m. The data across all sections and trips over a few consecutive weeks was used for
training with the last week in the data used for testing. We used data from Mon to Sat in a week as in Indian conditions Saturday is also a working day for a large majority of
people, while a previous study \cite{kumar:13} provides evidence of Sunday traffic being different from the rest of the days.
}

\subsection{Capturing Spatio-Temporal and Seasonal Correlations}
\label{sec:STSCorrelations}
{\bf Motivation:} One of the factors influencing  the section travel time $Z_n$ at section $n$, could be  its preceding section travel times. A justification for this can be as follows. 
The preceding section travel times  can  provide strong hints of propagating  congestions moving downstream along the route OR upcoming 
congestions (captured by above-average travel times) at the 
subsequent sections across which  travel times need to be predicted.  Patterns present in the historical data between 
section travel times can also be captured by this.

Further, in addition to the section travel times of the previously traversed sections of the currently plying bus, travel times experienced by 
previous buses (most recent) across the subsequent sections would be indicative of the most recent (real-time) traffic condition on the subsequent sections. This information can 
help  better estimate the potential travel times 
to be experienced by the current bus 
in the subsequent sections ahead. We indicate the associated section travel times by $Z_n^{pv}$ and the section entry times by $T_{n}^{e:pv}$, from the closest previous bus.  The section entry time $T_{n}^{e:pv}$
essentially tell us when the respective section travel times $Z_n^{pv}$ transpired. The section entry time information is important because larger the difference between  $T_{n}^{e:pv}$ and the current time $T_c$ (OR time of query), lesser the influence of $Z_n^{pv}$ on $Z_n$, the entity of interest.

In addition to the above real-time information, given strong weekly patterns in similar traffic data \cite{kumar:13} in general, we propose to exploit information from a historical trip from the previous week whose start time is closest to that of the current trip. We indicate associated section travel times by $Z_n^{pw}$ and section entry times by $T_{n}^{e:pw}$. This information would capture the weekly seasonal correlations in the data and would potentially enhance the predictive quality of the model.

\subsection{Dynamic Real-time Prediction Problem}
\label{sec:Prediction}
\begin{figure*}[!thbp]
\center
 \includegraphics[height=4.0in, width=7.0in]{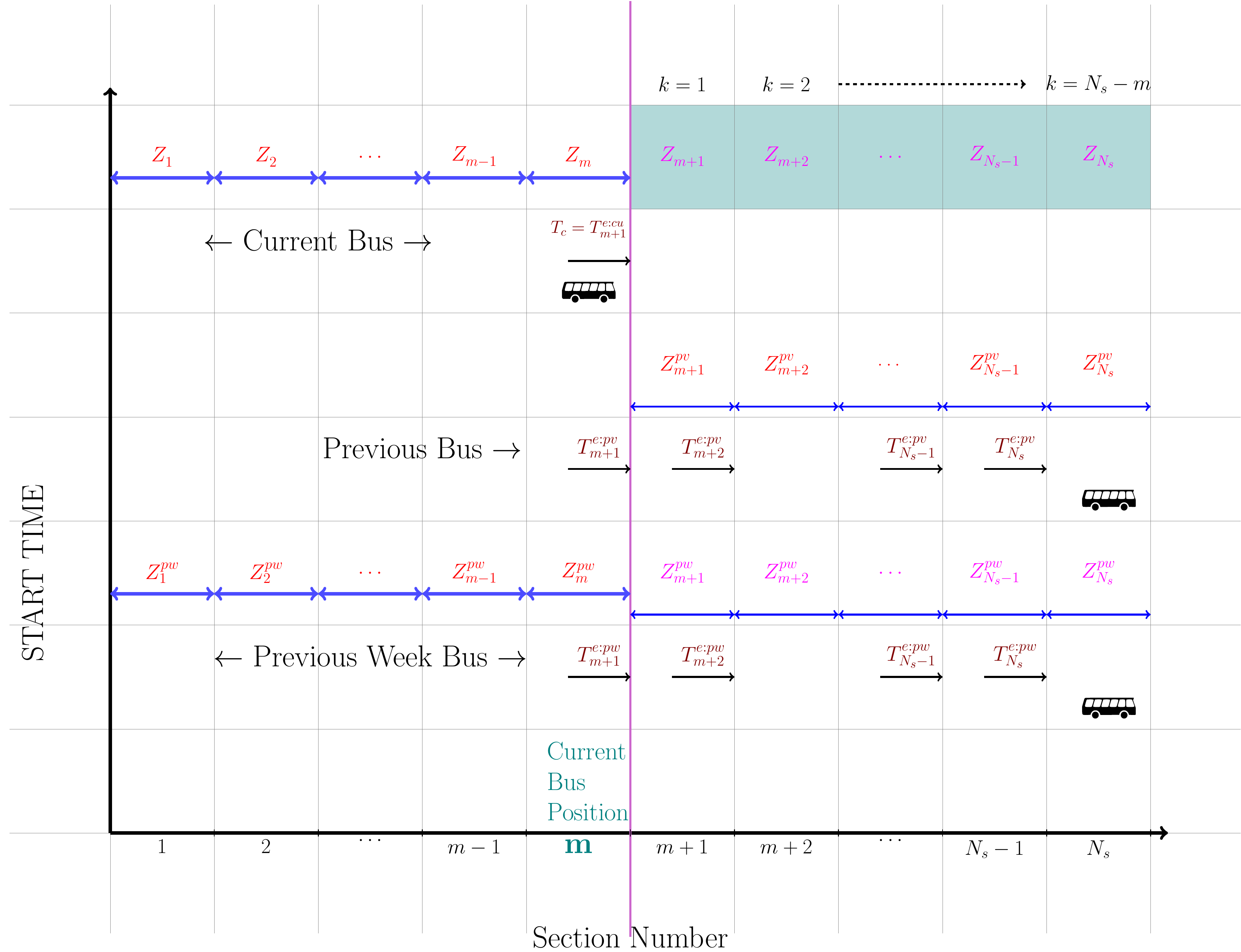}  
	\caption{Spatio-temporal and Seasonal Correlations pictorially. Current Bus Inputs (Spatial correlations), Previous Bus Inputs (Temporal Correlations), Previous Week Bus Inputs (Weekly Seasonal Correlations).}
\label{fig:FigCorrelations}
\vspace{-0.00in}
\end{figure*}

BATP, the dynamic real-time  prediction problem  can be formally stated as follows in view of the just described inputs potentially influencing 
prediction. 
Given (a)real-time position of current bus  
(say at the end of section $m$)  (b)current time (equivalent to $T^{e:cu}_{m+1}$, the current bus's entry time  into section $m+1$) 
(c)current bus's previous section travel times
(d)section travel times (ahead of section $m$) of the closest previous bus  and (e)section travel times (beyond section $m$) from  a historical trip from the previous week (but same weekday)  with the 
closest
trip start time to that of the current trip,   the task is  to estimate all section travel times beyond section $m$ of  the current bus.
\begin{table}[!thbp]
        \small
\caption{List of symbols}
        \center
        \begin{tabular}{|c|p{6.95cm}|}
                \hline
                 \textbf{Symbol} & \textbf{\begin{tabular}[c]{@{}c@{}}Descriptions\end{tabular}} \\ \hline
                 $n$ & Index for section number \\ \hline
                 $N_s$ & total number of sections in a route \\ \hline
                 $m$ & Section Index for the Current Bus Position \\ \hline
                 $T_c$ & Current Time OR time of prediction query. \\ \hline
                 $Z_n$ & travel time  across section $n$ of current bus.        \\ \hline
                 $Z_{n}^{pv}$ & travel time across section $n$ of closest previous bus. \\ \hline
                 $Z_{n}^{pw}$ & travel time across section $n$ of the closest bus from previous week, same weekday. \\ \hline
                $T^{e:cu}_{n}$  & entry time into section $n$ of current bus.         \\ \hline
                $T^{e:pv}_{n}$  & entry time into section $n$ of closest previous bus.        \\ \hline
                $T^{e:pw}_{n}$  & entry time into section $n$ of (closest) previous week bus, same weekday.        \\ \hline
                $K$ & Number of steps in the Decoder (Fig.~\ref{fig:EDArchitecture}, Fig.~\ref{fig:BidirDec}), also  equal to $(N_s - m)$. \\ \hline
                $F$ & Predictive (Regression) Function learnt by the ED architecture. \\ \hline
\end{tabular}
        \label{tab:SymbolTab1}

\end{table}

Fig.~\ref{fig:FigCorrelations} gives a clear pictorial spatial layout of all the relevant input entities that influence prediction and the associated target
variables of interest. $m$ denotes the section index of the current bus position (top bus in Fig.~\ref{fig:FigCorrelations}), while $N_s$ denotes the total number of sections. 
$Z_1,Z_2,\dots Z_m$ denote the traversed section travel times of the current bus. 
$Z_{m+1},Z_{m+2}\dots Z_{N_s}$ are the target variables (marked with a shade). 
They denote the section travel times of current bus across the subsequent sections, which need to be predicted. Similarly, $Z_{m+1}^{pv},Z_{m+2}^{pv},\dots Z_{N_s}^{pv}$, denote section travel times across subsequent sections of the closest previous bus (middle bus in Fig.~\ref{fig:FigCorrelations}), while $T_{m+1}^{e:pv},T_{m+2}^{e:pv},\dots,T_{N_s-1}^{e:pv},T_{N_s}^{e:pv}$ 
denotes the associated section entry times. The bottom bus in Fig.~\ref{fig:FigCorrelations} denotes the closest trip from the previous week, same weekday and its associated 
section travel and entry times  along the entire route are clearly indicated. 
  The main symbols employed in this paper have been summarized in Table~\ref{tab:SymbolTab1}.

Given current position $m$ and current time $T_c$, which is also the entry time of the bus into section $m+1$ (i.e. $T^{e:cu}_{m+1}$), we wish to 
	learn an input-output function $F(.)$ of the below form. 
\begin{align}
	&\Big(Z_{m+1},Z_{m+2},\dots\dots\dots,Z_{N_s - 1}, Z_{N_s} \Big) = \nonumber \\
	&{F\left(m,T_c,\underbrace{Z_{m},Z_{m-1},\dots\dots\dots,Z_2,Z_1,} \right.} \nonumber \\
	&{\dashuline{Z_{m+1}^{pv},Z_{m+2}^{pv},\dots,Z_{N_s-1}^{pv},Z_{N_s}^{pv},}
	\dashuline{T_{m+1}^{e:pv},T_{m+2}^{e:pv},\dots,T_{N_s-1}^{e:pv},T_{N_s}^{e:pv},}}\nonumber \\
	&{\underline{Z_{m+1}^{pw},Z_{m+2}^{pw},\dots,Z_{N_s-1}^{pw},Z_{N_s}^{pw},}
\underline{T_{m+1}^{e:pw},T_{m+2}^{e:pw},\dots,T_{N_s-1}^{e:pw},T_{N_s}^{e:pw},}}\nonumber \\
        & {\left. \underline{Z_{m}^{pw},Z_{m-1}^{pw},\dots\dots\dots,Z_2^{pw},Z_1^{pw},}  \right) }
    \label{eq:SpatialMultiStep}
\end{align}
The inputs have been grouped into $3$ categories (based on the style of underlining). The first category corresponds to the current bus
	information, namely its current position $m$, its section travel times along traversed sections (correspond to spatial correlation). The second category is the section travel
	and entry times from the closest previous bus across all subsequent sections (correspond to temporal correlations). The third category  includes section travel time from 
	all sections of the closest previous week trip, while its section entry times from only the subsequent sections (correspond to weekly seasonal correlations). Note we don't use the previous section travel
times from the previous trip as the current trip travel times across the previous sections are more recent.

\subsection{Motivating the proposed ED Variant Architecture}
\label{sec:EDArch}
\begin{figure*}[!thbp]
\center
 \includegraphics[height=4.0in, width=7.0in]{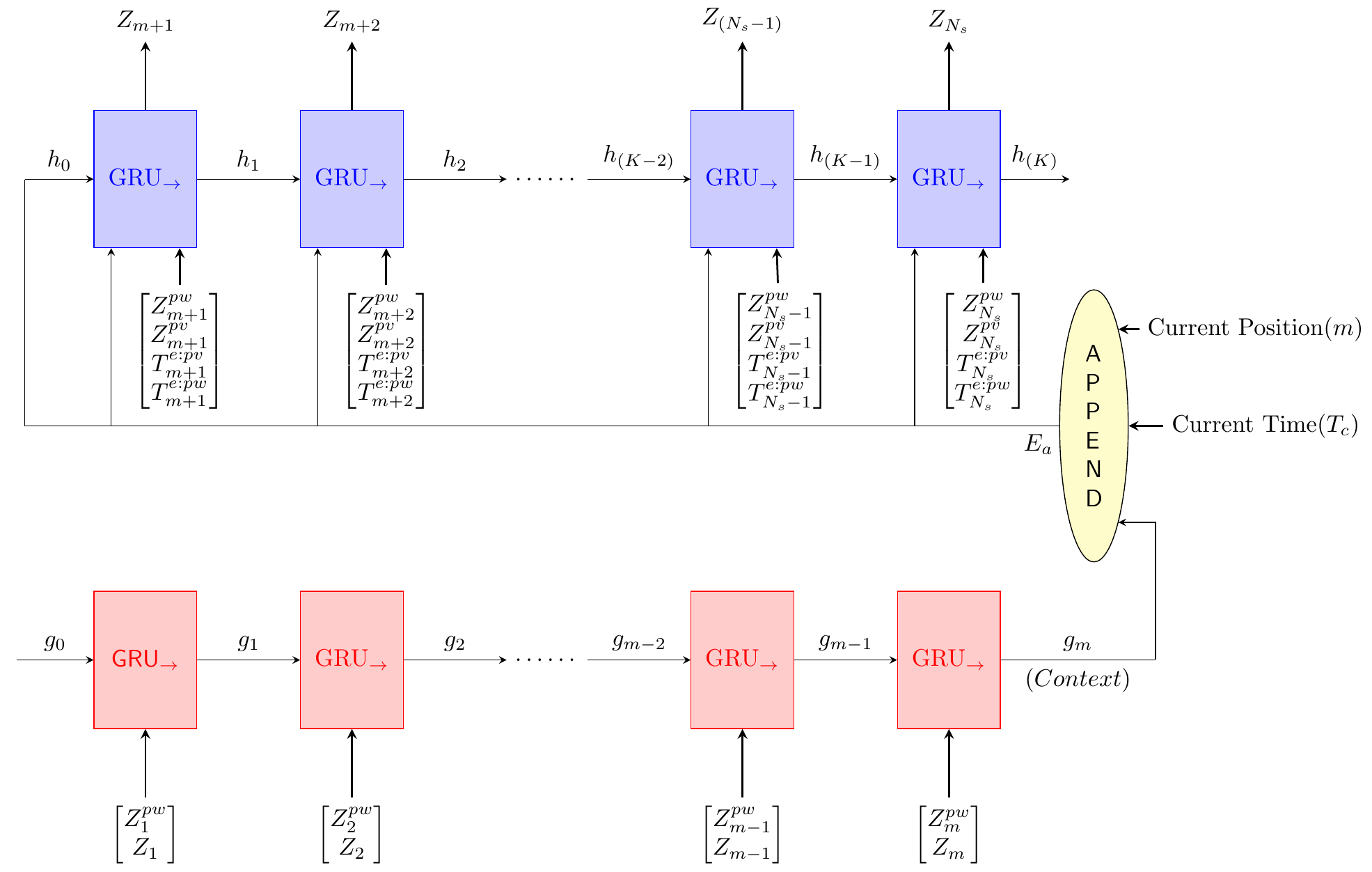}  
	\caption{Proposed ED Architecture with a Unidirectional Decoder. $K=(N_s - m)$}
\label{fig:EDArchitecture}
\vspace{-0.00in}
\end{figure*}

{\bf Traditional ED:} The ED architecture was originally employed for variable length pairs of input-output sequences. The original Seq2Seq idea was to employ two distinct RNN layers referred to as an Encoder (colored in red) and Decoder (colored in blue) respectively, as shown in Fig.~\ref{fig:EDArchitecture}. The first RNN layer takes the input sequence as input and the state computed at the last step of the unfolded structure is fed as initial state of the second layer. This state can also be fed as an input at every step of the second layer (decoder). The output sequence whose length is independent of the input sequence is the output of the unfolded decoder. 

In the earliest machine translation context, the input could correspond to a sentence from a
particular language, while the output could be its translation in another language. Each word essentially comes from a categorical space and
one needs sophisticated word2vec \cite{Pennington14,Mikolov13} representations before feeding the transcribed words into the RNN.  In our setting the raw data is already
a real number and hence can be fed directly into the RNNs.

{\em Our prediction problem (BATP) can be viewed as a spatial multi-step prediction problem,} where at $i^{th}$ step we can either predict (i)the travel time across the
$(m+i)^{th}$ segment OR (ii)travel time to reach end of $(m+i)^{th}$ segment. We stick to the former in this paper.

\begin{remark}
The important point to note from eqn.~\ref{eq:SpatialMultiStep} is that both the input length  and the output length of the $F(.)$ function changes with $m$, 
the current bus Position. For instance, the output sequence length for a bus position $m$ is ($N_s - m$). Similarly the contribution to the input sequence length from 
the current bus real-time inputs is $m$.   This means the training data for BATP actually has a clear variable length input-output nature.
Given the variable length Seq2Seq mapping ability of the ED framework, this can  aptly be utilized for BATP with variable length input and outputs. 
\end{remark}

So the spatial multi-step (target) prediction problem 
we encounter has both variable-length inputs and  variable-length vector valued targets,  with vector size equal to the number of sections ahead ($N_s - m$). 
We rewrite eqn.~(\ref{eq:SpatialMultiStep})  by reorganizing its inputs as follows
which aids us in clearly associating the inputs and outputs of
the regression function $F(.)$ to the proposed
ED architecture.
\begin{align}
	&\Big(Z_{m+1},Z_{m+2},\dots\dots\dots,Z_{N_s - 1}, Z_{N_s} \Big) = \nonumber \\
	&{F\left(m,T_c,\underbrace{(Z_{m},Z_{m}^{pw}),(Z_{m-1},Z_{m-1}^{pw})\dots,(Z_2,Z_2^{pw}),(Z_1,Z_1^{pw}),} \right.} \nonumber \\
	&{\underline{(Z_{m+1}^{pv},Z_{m+1}^{pw},T_{m+1}^{e:pv},T_{m+1}^{e:pw}),(Z_{m+2}^{pv},Z_{m+2}^{pw},T_{m+2}^{e:pv},T_{m+2}^{e:pw})\dots,}
	}\nonumber \\
	&{\left.\underline{\dots,(Z_{N_s-1}^{pv},Z_{N_s-1}^{pw},T_{N_s-1}^{e:pv},T_{N_s-1}^{e:pw}),(Z_{N_s}^{pv},Z_{N_s}^{pw},T_{N_s}^{e:pv},T_{N_s}^{e:pw})}\right)
} 
    \label{eq:EncDecMulti-StepGen}
\end{align}
All the previous section travel times from the current bus and the previous week trip have been grouped. These two pairs of inputs 
are fed as encoder inputs unfolded up to $m$ steps (Fig.~\ref{fig:EDArchitecture}), which makes the input dimension vary with $m$. In the second group, we bring together section travel
times and entry times of the {\em closest} previous bus and an appropriate previous week trip. These inputs in pairs of four are fed as additional inputs
at each step of the decoder, where the decoder is unfolded into  $N_s - m$ steps, making the output dimension also vary with $m$. The intuition is that not just the subsequent section travel 
time, but also the time at which the traversal happened (entry time into that section) has an influence. Closer the entry time of the previous bus
to the current bus's likely entry time, higher is its influence. The idea is that the model learnt would capture this influence from the data. In this architecture, the current bus's likely entry time into section
$m+i$ would be inherently represented in the hidden state $h_i$.
\begin{remark}
 The GPS/AVL based travel-time data is naturally discrete in the spatial dimension, because we have travel-time information available either across sections OR between two consecutive bus-stops as described earlier. On the other hand this data is naturally continuous in temporal dimension as trips  can happen anytime during the day. 
	So it makes more sense to map the (discrete) spatial aspect of the problem to the discrete sequential aspect of the Seq2Seq (or ED) framework. Further, real-time BATP needs to predict travel times across all subsequent sections (from the current section). This spatial and sequential multi-step target aspect of BATP can be exactly mapped to 
	the multi-step sequential decoder outputs. The temporal correlations coming from the previous bus are fed as decoder inputs in a space synchronized fashion, while the weekly seasonal historical inputs are clearly split between the encoder and decoder inputs without any redundancy. Our ED framework hence also avoids discretizing time as in 
\cite{Petersen19,Ran19,Wu20}.
\end{remark}

\subsection{ED Variant with Bidirectional Decoder }
\subsubsection{\bf Gating Unit in the architecture}
Between  LSTM~\cite{Hochreiter97} and GRU~\cite{Chung14} RNN unit choices, both of which have a gating mechanism to check vanishing gradients and have persistent memory, we choose GRU in this paper.   GRU unit  is currently
very popular in various sequence prediction
applications \cite{Gupta17,Ravanelli18,Che16,Nicole20}  as the building block for RNNs. Moreover, compared to LSTM it has lesser gates, in turn leading to lesser weight parameters.
The hidden state of a single hidden layer (simpler) plain RNN unit  can be written as
\begin{equation}
        h_i = \sigma(W^{h} h_{i-1} + W^{u} u_i)
\end{equation}
where $W^{h}$, $W^{u}$ are the weight matrices associated with the state at the previous step $h_{i-1}$ and the current input ($u_i$) respectively,
$\sigma(.)$ denotes sigma function.
 The hidden state of a GRU based cell (for one layer) can be computed as follows.
\begin{eqnarray}
        z_i & = &\sigma(W^z u_i + U^z h_{i-1})  \label{eq:UpdateGate} \\
        r_i & = & \sigma(W^r u_i + U^r h_{i-1}) \\
        \tilde{h}_i & = &tanh(r_i \circ U h_{i-1}  + W u_i) \\
        h_{i} & = & z_i \circ h_{i-1} + (1 - z_i)\circ \tilde{h}_i \label{eq:StateUpdate}
\end{eqnarray}
where $z_i$ is update gate vector and $r_i$ is the reset gate vector.  We obtain the plain RNN unit if these two gates were absent,. $\tilde{h}_i$ is
the new memory (summary of all inputs so far) which is a function of $u_i$ and $h_{i-1}$, the previous hidden state. The reset signal controls the
influence of the previous state on the new memory. The final current hidden state is a convex combination (controlled by $z_i$) of the new memory and the memory at the previous
step, $h_{i-1}$.   We use back-propagation through time (BPTT) to train all associated weights $W^z$, $W^r$, $W$, $U^z$, $U^r$, $U$.

\subsubsection{\bf Bidirectional Layer at the Decoder}
\label{sec:Bidir}

\begin{figure*}[!htbp]
\center
 \includegraphics[height=2.4in, width=7.0in]{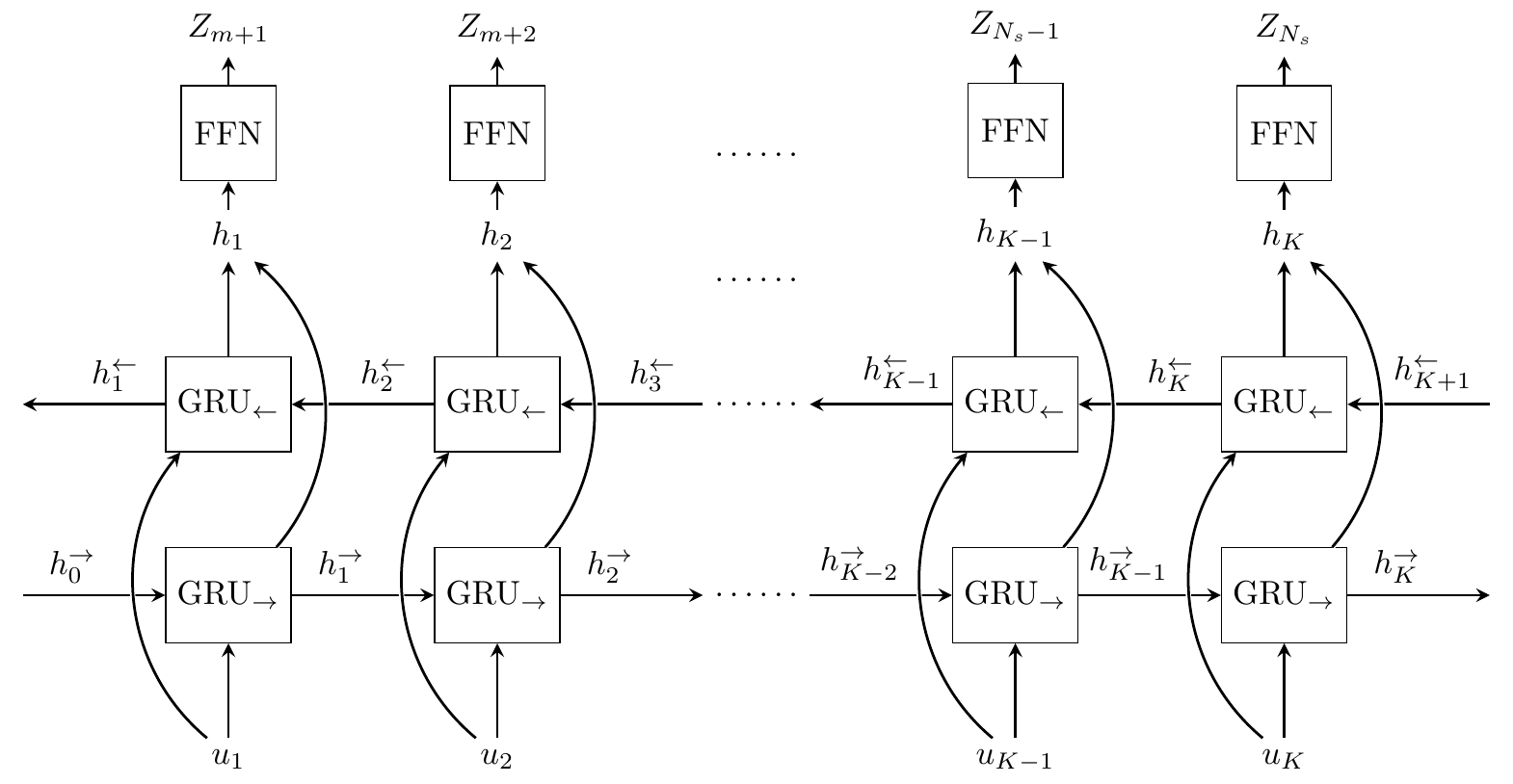}  
	\caption{Bidirectional Decoder. Please note $K=(N_s - m)$. $h_{K+1}^\leftarrow$ and $h_0^\rightarrow$ are set to  $E_a$, output of the
        append block in Fig.~\ref{fig:EDArchitecture}.}
\label{fig:BidirDec}
\vspace{-0.00in}
\end{figure*}

We further propose to use a bidirectional layer at the decoder. The motivation for this comes primarily from this application. To predict $Z_{m+k}$,
the  unidirectional architecture as in fig.~\ref{fig:EDArchitecture} considers the previous bus's travel times up to section $m+k$ only. However, the
previous bus's travel times in subsequent sections beyond $m+k$ can provide indicators of recent congestions further down in the route. {\em These recent
congestions can in turn potentially propagate backward in space and strongly influence the travel time of the current bus at section $(m+k)$.} To
capture such eventualities, we use a bidirectional layer as given in fig.~\ref{fig:BidirDec}.

In the GRU-cell defining equations (eqn.~(\ref{eq:UpdateGate})-(\ref{eq:StateUpdate})), the state-update essentially follows the below equation
 \begin{equation}
 h_i^\rightarrow=f_1(h_{i-1}^\rightarrow,u_i)
 \end{equation}
 where for each time-step the state information flows from left to
right. In a bi-directional setting, we have an additional state-vector $h_i^\leftarrow$ and map $f_2$, with an update with reverse information flow from right to left as
follows.
 \begin{equation}
h_i^\leftarrow=f_2(h_{i+1}^\leftarrow,u_i)
 \end{equation}

   State at step $i$, $h_i$, is a concatenation $[h_i^\rightarrow, h_i^\leftarrow]$. Note that $f_2$ like $f_1$ is governed by the same GRU-cell defining equations
(eqn.~(\ref{eq:UpdateGate})-(\ref{eq:StateUpdate})) with possibly different weight values. Each input $u_i$ in Fig.~\ref{fig:BidirDec} is
actually the concatenation of all inputs at the $i^{th}$ (each) step of the decoder in Fig.~\ref{fig:EDArchitecture}. In particular,
\begin{equation}
        u_i = \left[ Z_{m+i}^{pw} \,\, Z_{m+i}^{pv} \,\,  T_{m+i}^{e:pv} \,\,T_{m+i}^{e:pw} \,\, E_a \right]
\end{equation}
where $E_a$ is output of the append block in Fig.~\ref{fig:EDArchitecture}.
Also note initial states
($h_{K+1}^\leftarrow$) and ($h_0^\rightarrow$) are equal and initialized to $E_a$. Finally, $Z_{m+i} = f_3(h_i)$, where $f_3$ is a feed-forward map.

\subsection{\bf Training Data Preparation}
 Preparation of the training data for training our proposed ED models involves some extra effort. Specifically, we need inputs from the closest previous bus  of the same day. We cannot just pick 
 the closest previous bus based on the start time of the trip. The closest previous bus at every subsequent section can be potentially different. Bus bunching in the data can make the scenario pretty complicated. For simplicity, let us assume there is no bus bunching. Then  the immediate previous bus, at current time, $T_c$ may be a few section ahead. For those few sections, the closest previous bus 
 section traversal will come from this bus. But for sections beyond the current position of this immediate previous bus, we will have to look at buses which started before this immediate previous bus.     Hence for every subsequent section, we need to search for the most recent bus (just before $T_c$) which traversed that section. We need to compare $T_c$ with the entry time into the section from each of the trips and pick that trip whose entry time is before $T_c$ and  closest  to $T_c$. Recall this will form as a part of the input to the decoder. Computing this field of the input for each subsequent section of a training example will be $\scr{O}(N_{tr})$, where $N_{tr}$ is the no. of trips on that day. In a general bus bunching scenario, one needs to search over all trips of that day. For each $m$, there are $N_s - m$ such subsequent sections.  So for each trip, one needs to perform this $\scr{O}(N_{tr})$ search $N_s * (N_s + 1)/2$ times.   This furher needs to be done across all trips and all training days. If $N_{day}$ denotes the number of days of training data and if $N_{tr}$ denotes the average no. of trips per day, then the complexity of data preparation now will be $\scr{O}(N_{day} *N_{tr}^2 * N_s^2)$.    Since this is a one-time pre-training step, its not a problem.      Even if one wants to retrain with new data later, this effort need not be repeated on old data.

\section{Results}
\label{sec:Results}
{\bf Data:} We tested and bench-marked our method on one bus route (Route 399)  from Delhi, India in detail. A pictorial view of the route is provided in App.~\ref{app:Route}. The route was uniformly segmented into sections of length
$800$m resulting in a total of $34$ segments.  The section width was chosen keeping in mind (a)the actual number of bus-stops and (b)the number of resulting sections ($N_s$). 
Having too many
sections can be challenging on the model for long sub-route prediction, while long section length would mean poor predictions on short sub-route predictions.
For every week, data from Mon to Sat was considered for training. Saturday data was also included as it also happens to be a working day  for a large segment of
people in Indian conditions. 
Further, Sunday was excluded as a previous study \cite{kumar:13} under similar conditions provides evidence of Sunday traffic being distinct. 
We bench-marked on data collected from all trips over two months (Sep-Oct $2019$). We filtered the noise from the raw GPS
measurements before using it for training our models. Filtering was performed based on simple route projection by exploiting the known route information.   
For training, we used the first $7$ weeks of data, while the  final ($8^{th}$) week's data  was kept aside for testing.  

Evaluation was carried out based on $2$ complementary metrics: Mean Absolute Error (MAE) and Mean Absolute Percentage Error (MAPE). {\em Percentage error is Absolute Error divided by 
true prediction expressed in percentage.}  While MAE provides  a user 
understandable clock time difference in seconds, MAPE is a scale independent metric.  
Generally, accurancy of short route prediction is crucial for commuters planning their arrival on time to the bus-stop. Generally, such commuters will be planning to reach their closest bus-stops on time. On the other hand, accurate mid and long route predictions is crucial for commuters deciding whether to board the bus or take an alternative mode of transport.

\subsection{Bench-marking Details} 
{\bf Proposed Approaches:} We denote our proposed ED approaches as EDU (unidirectional decoder) and EDB (bidirectional decoder).   The bidirectional model version can lead to many more parameters in the decoder (in comparison to the unidirectional version) for a similar number of hidden 
nodes in the GRU cell. For consistency in the number of learnable parameters, the number of hidden nodes in the GRU-cell of the bidirectional decoder is kept lower so that 
the overall number  of parameters in the bidirectional decoder matches that of the unidirectional decoder. 

{\bf Baselines:} In this paper, we benchmark  the proposed  methods  (with or without bidirectional layer at the decoder) against $4$ recent state-of-art baselines all of which capturing spatio-temporal (ST) correlations in a distinct way: 
\begin{itemize}
	\item (a)LNKF \cite{Achar19a},   which learns spatio-temporal correlations (post a log transformation) using linear statistical models first  followed by a linear kalman filter prediction approach
	\item (b)SVKF \cite{Achar19b}, which learns spatio-temporal correlations using nonlinear support-vector approximations followed by a extended kalman filter prediction  approach.
\item (c)DpAR\cite{Flunkert17}
many-to-many architecture inspired from
		\cite{Flunkert17}  (an RNN approach for sequential (time-series) prediction with possible exogenous inputs).  The current bus's section travel time from previous section (akin to previous sequential input of DeepAR)  and
previous bus's  section travel time from current section (like the exogenous input of 
DeepAR) are fed as inputs to predict the current bus's current section travel time at each time step. This baseline (indicated as DpAR from now on) captures
ST correlations in a distinct way.
\item (d)CLSTM \cite{Petersen19} a current interesting spatio-temporal approach utilizing Seq2Seq to capture temporal correlations and Convolutional layer to capture spatial correlations.
\end{itemize}

BPTT was used to train all the deep learning algorithms namely DpAR, CLSTM, EDU and EDB. All these methods use Adam optimizer with batch size=32. To tackle over-fitting, early stopping was used for 
DpAR, EDU and EDB, while the default batch normalization (as prescribed in \cite{Petersen19}) was used for CLSTM.  For LNKF training (model building), we used a least squares regression on the log transformed inputs. For SVKF training, support vector regression with a Gaussian kernel was used, while a grid search was employed to narrow down on the  hyper-parameters, $\epsilon$ and C.

Model training (either for the proposed approaches OR the baselines considered) is carried out separately in an offline fashion using historical data of the previous few months. 
Here we chose two previous months. The learnt model in conjunction with real-time inputs from the current bus and closest previous bus (and a historical seasonal trip from previous week) is now employed for dynamic arrival time prediction in real-time.

{\bf Justification of baselines:} Please note LNKF and SVKF are recent learning-based spatio-temporal approaches which also capture the spatial sequential aspect of BATP in their predictive models 
like our approach, but by performing a  Kalman-filter based spatial multi-step prediction. SVKF in particular is a recent interesting nonlinear approach using support vector approximations for model building. CLSTM and DpAR are two recent DL approaches which make them a natural choice for baselining. Owing to the following reasons, we do not benchmark our method against some of the other 
recent approaches like \cite{KDD20},\cite{Vignesh20},\cite{Dhivya20} discussed in related work. 
BusTr \cite{KDD20} considers run-time and dwell time separately, while takes very different inputs like speed forecasts, location embeddings etc. 
which are very hard to procure and compute for the data and route on which we bench-marked. 
The performance of the TD approach of \cite{Vignesh20} on the other hand was very poor on short and mid-length sub-routes. 
CLSTM is a nonlinear model which captures temporal correlations on lines similar to \cite{Dhivya20} by segmenting the time axis. While \cite{Dhivya20}  captures linear temporal 
correlations only, CLSTM additionally captures spatial correlations also via a convolutional layer making it a stronger baseline compared to \cite{Dhivya20}.  
Overall our choice of baseline methods is current and enables a  
diverse comparison.

{\bf Assessing significance of Error differences statistically:}
We have conducted a Z-test based significance assessment (across all relevant experiments) under MAPE/MAE differences
(EDB vs EDU/Baseline) with a significance level of 0.1 to reject the null hypothesis. The mean error metric is evaluated by averaging
at-least  30 samples in all experiments and hence a Z-test is sufficient. 

{\bf Model Building Details:} Recall that Fig.~\ref{fig:EDArchitecture} takes current position ($m$) also as input and builds a unified model.  One issue with the unified model is that the section
number has to be one-hot encoded with a large number of binary inputs (equal to the number of sections, about $34$ here). This can lead to huge number (quadratic increase) of weights while the ability to generalize across a large number of sections (current positions) can be challenging. Empirically we observe a poor performance in this approach. 
A possible intuitive explanation could be that the unified model has to generalize across all $m$  
while also handling a huge variation in encoder/decoder unfolded lengths for different $m$. To tackle this, we build a common model across every $5$ consecutive sections. In this way, we avail the variable (input-output) length feature of ED while the weights introduced due to one-hot encoding of the position is also under control. 
We have built models starting from section $3$, where the first model is trained with input-output pairs for $m$ ranging from $3-7$. In this way we build 6 models to cover the entire route, where each model can generalize across $5$ consecutive sections for current bus position ($m$). 

\subsection{Two-step (section) ahead Prediction (Short Routes)}
\label{sec:OneStep}
\begin{figure}[!thbp]
	\centering
	\subfigure[MAPE]{\includegraphics[width=3.4in,height=2.1in]{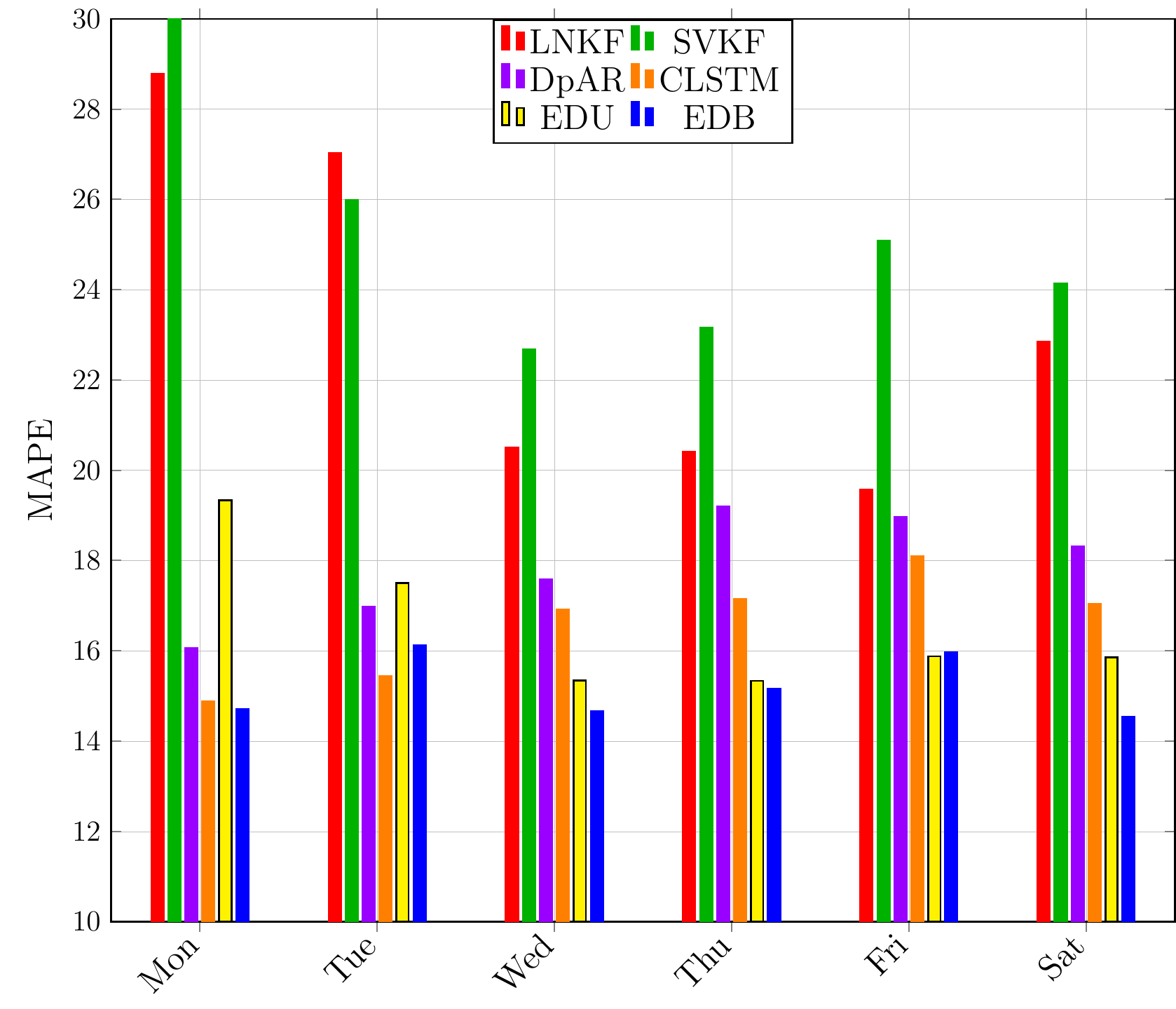}  \label{fig:TwoStepMAPE} }
\subfigure[MAE]{\includegraphics[width=3.4in,height=2.1in]{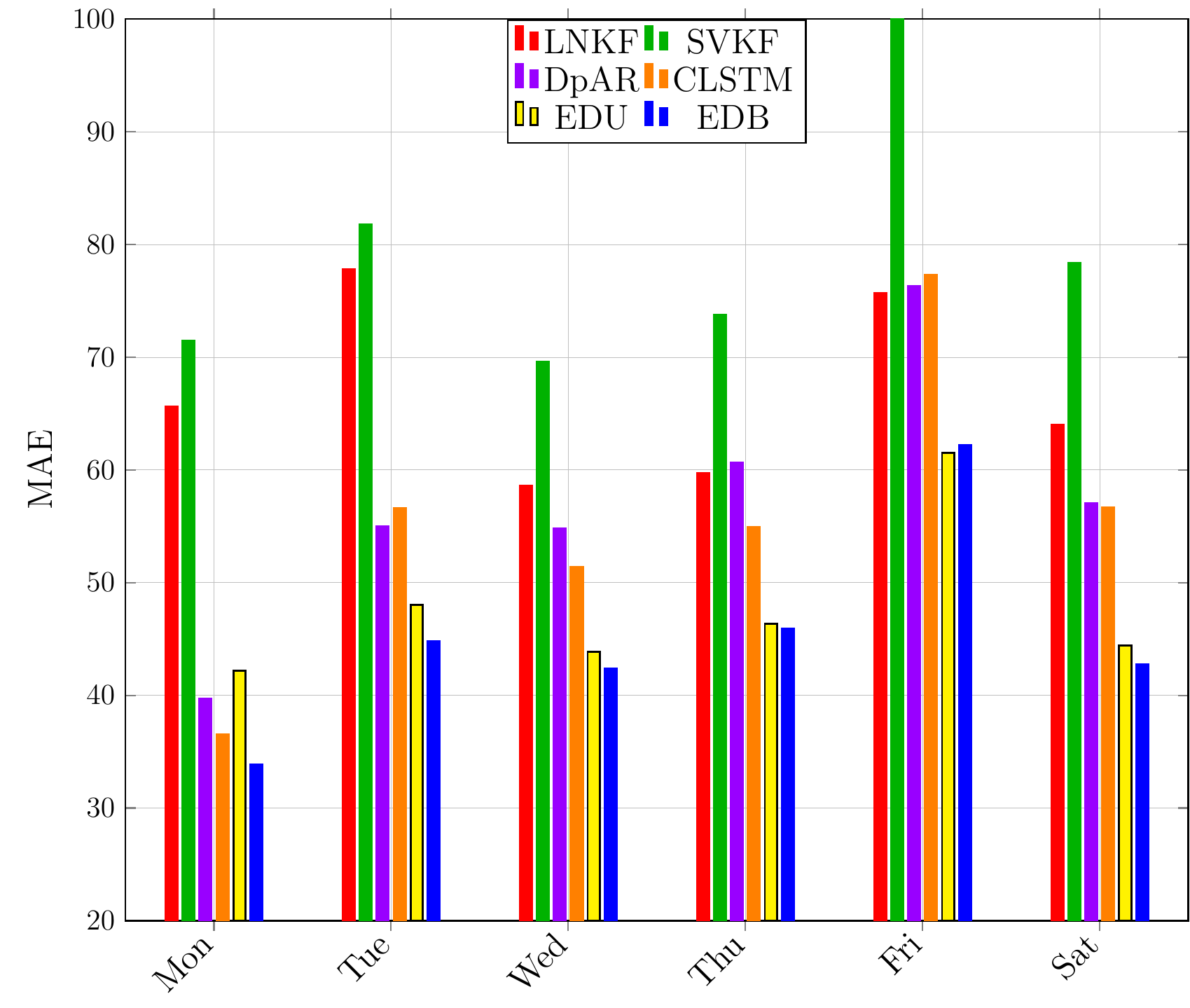} \label{fig:TwoStepMAE} }
	\caption{Two-Step Ahead MAPE and MAE Comparison at a Day level}
	\label{fig:TwoStep}
\end{figure}
\vspace{-0.02in}

{\bf Day Level Comparison:} We compare two-step ahead predictions (for short routes) of EDU and EDB with all existing methods at a day level.  Fig.~\ref{fig:TwoStep} 
illustrates  that EDB, our bidirectional proposed variant outperforms EDU and most of the state-of-art baselines  consistently 
 at a $2$-step level. 
Fig.~\ref{fig:TwoStepMAPE} and Fig.~\ref{fig:TwoStepMAE} provide the comparisons in terms of MAPE and MAE respectively. 
 In particular, the MAE prediction improvements of EDB over CLSTM are statistically significant (based on Z-test as described earlier) on $4$ days, while the two methods have 
 similar performance (statistically) on the remaining two days. Compared to DpAR, EDB's improvements are statistically significant on $4$ days while performance is similar on the remaining two days. 
 EDB performs better (statistically) on all days compared to EDU, LNKF, and SVKF. The MAPE prediction improvements of EDB are similar in trend to the MAE case. A slight exception here is in comparison to DpAR, where  EDB's predictions are better  on $3$ days, similar on $1$ day, while inferior to DpAR on $2$ days. 
In terms of the best-case improvements, EDB performs the best  with an advantage up to 10.21\%,  12.54\%, 4.03\%, 7.78\%
  and 8.93\% over LNKF, SVKF, DpAR, CLSTM   and EDU approaches respectively. 
Similarly in terms of MAE,  EDB performs the best  with an advantage up to 
29.87s,  53.88s, 30.81s, 34.11s  and 19.81s over  LNKF,  SVKF, DpAR, CLSTM   and EDU approaches respectively.
Overall, we observe   reasonable  improvements from our method, EDB,  based on both metrics.

\subsection{Multi-Step Prediction} 

\begin{figure}[!htbp]
	\centering
\subfigure[Bus current position at section $5$ and section
	$30$.]{\includegraphics[width=3.5in,height=2.6in]{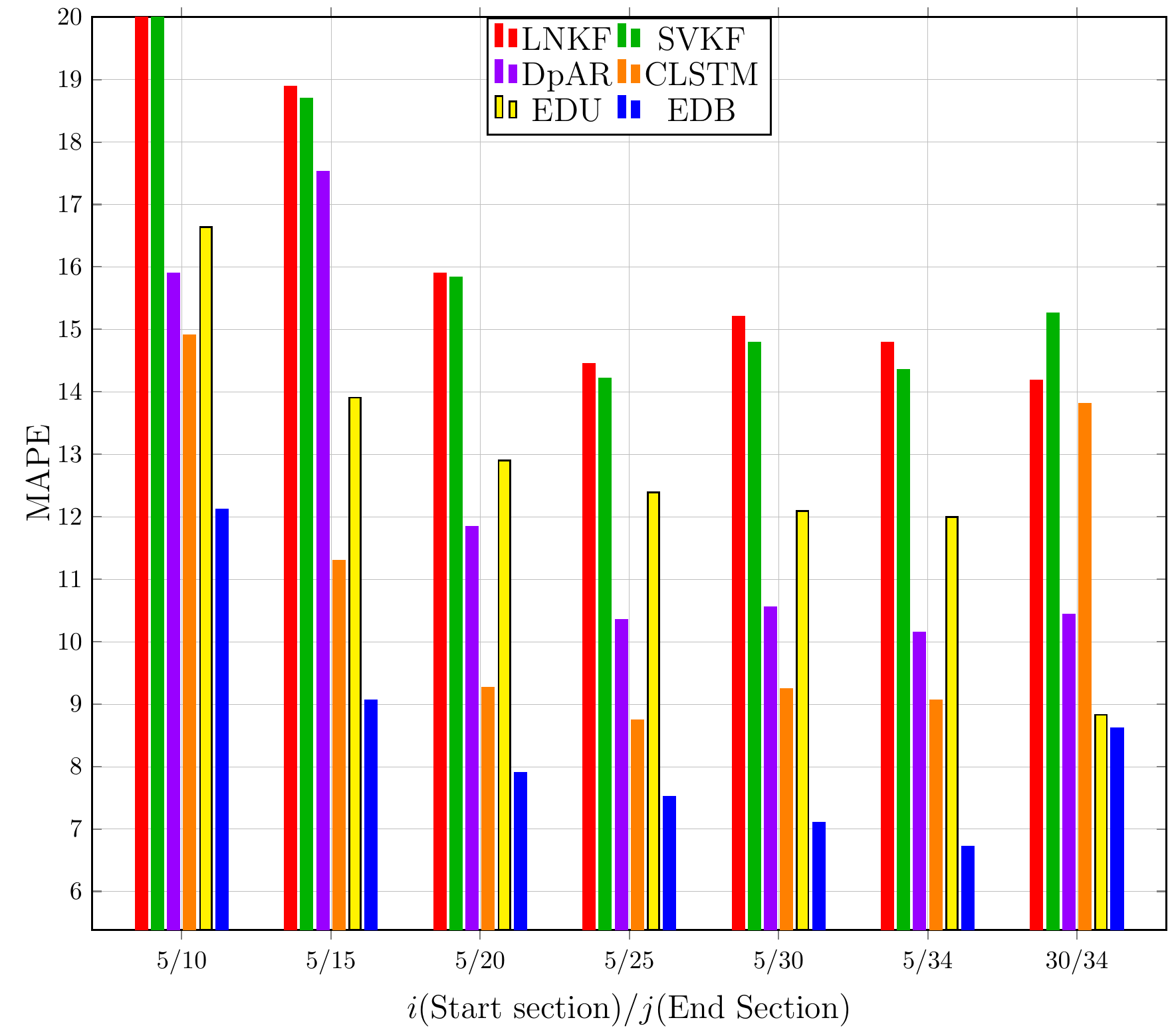} \label{fig:MAPESec530} }
\subfigure[Bus current position at section $10$ and section $25$.]{\includegraphics[width=3.5in,height=2.6in]{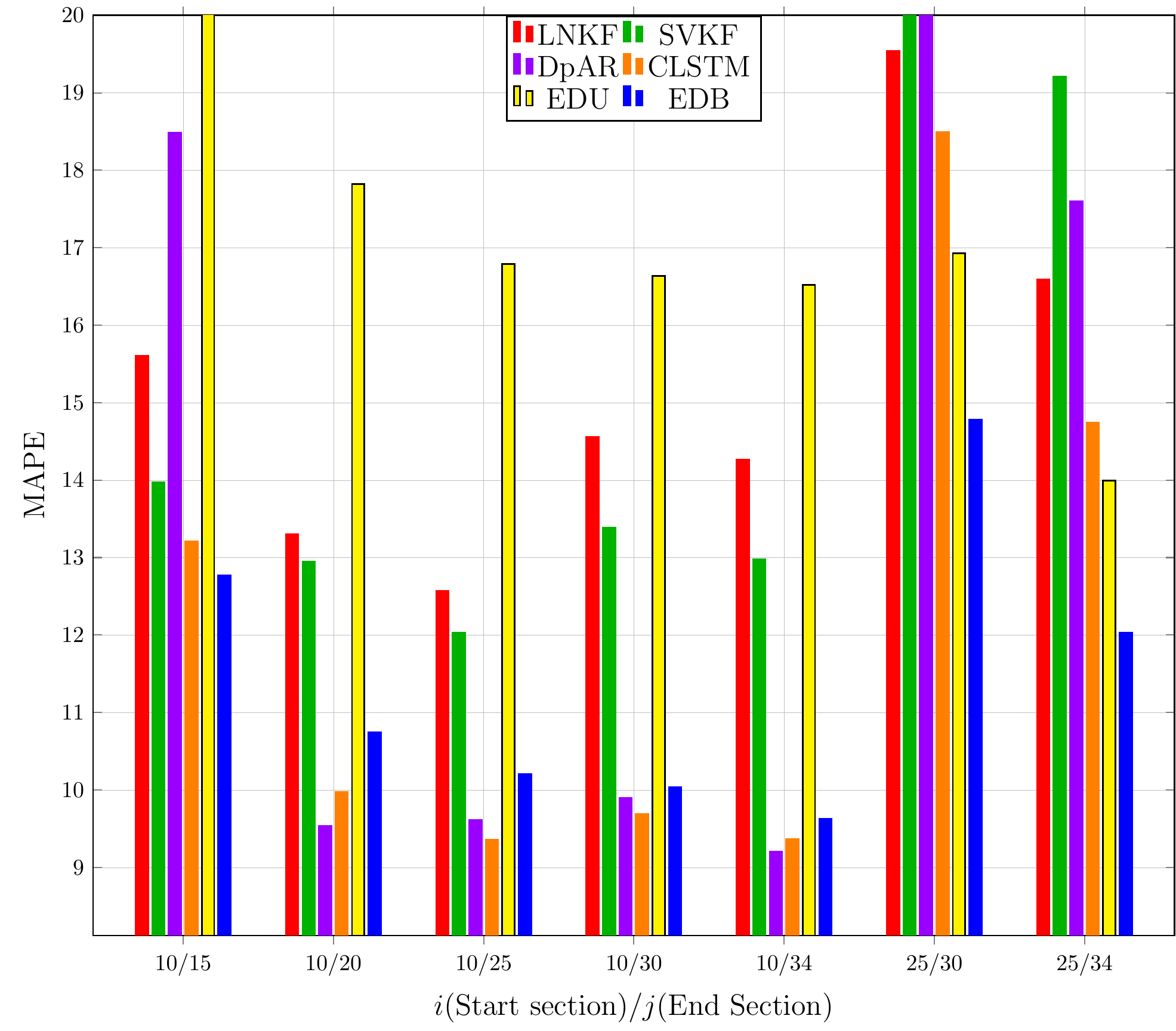}
\label{fig:MAPESec1025} }
\subfigure[Bus current position at section $15$ and section $20$.]{\includegraphics[width=3.5in,height=2.6in]{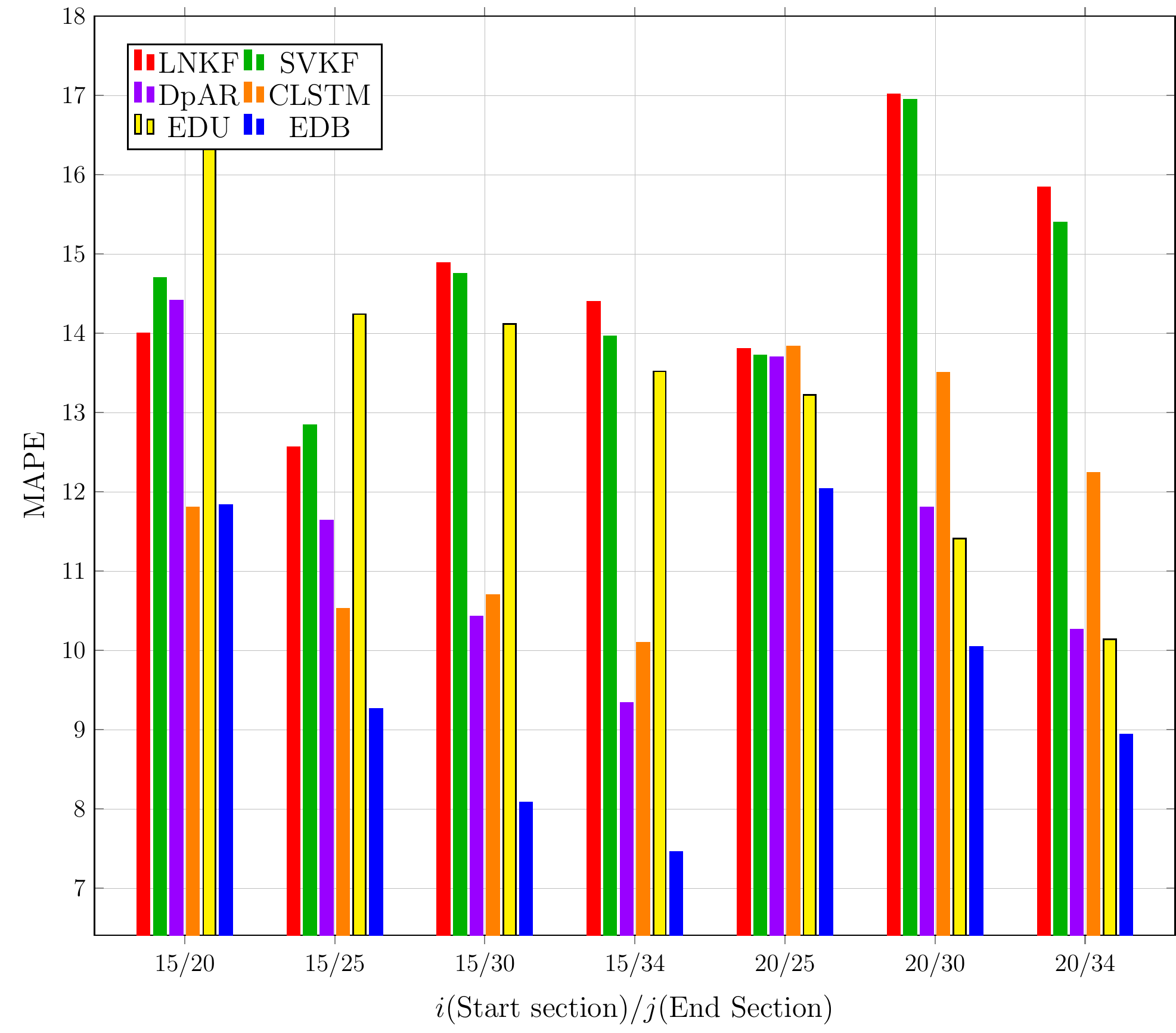}
\label{fig:MAPESec1520} }
	\caption{Multi-Step Ahead MAPE at various start/current bus positions (again chosen in steps of 5) along the route.}
	\label{fig:MultistepMAPE}
\end{figure}
\begin{figure}[!htbp]
	\centering
\subfigure[Bus current position at section $5$ and section $30$.]{\includegraphics[width=3.5in,height=2.6in]{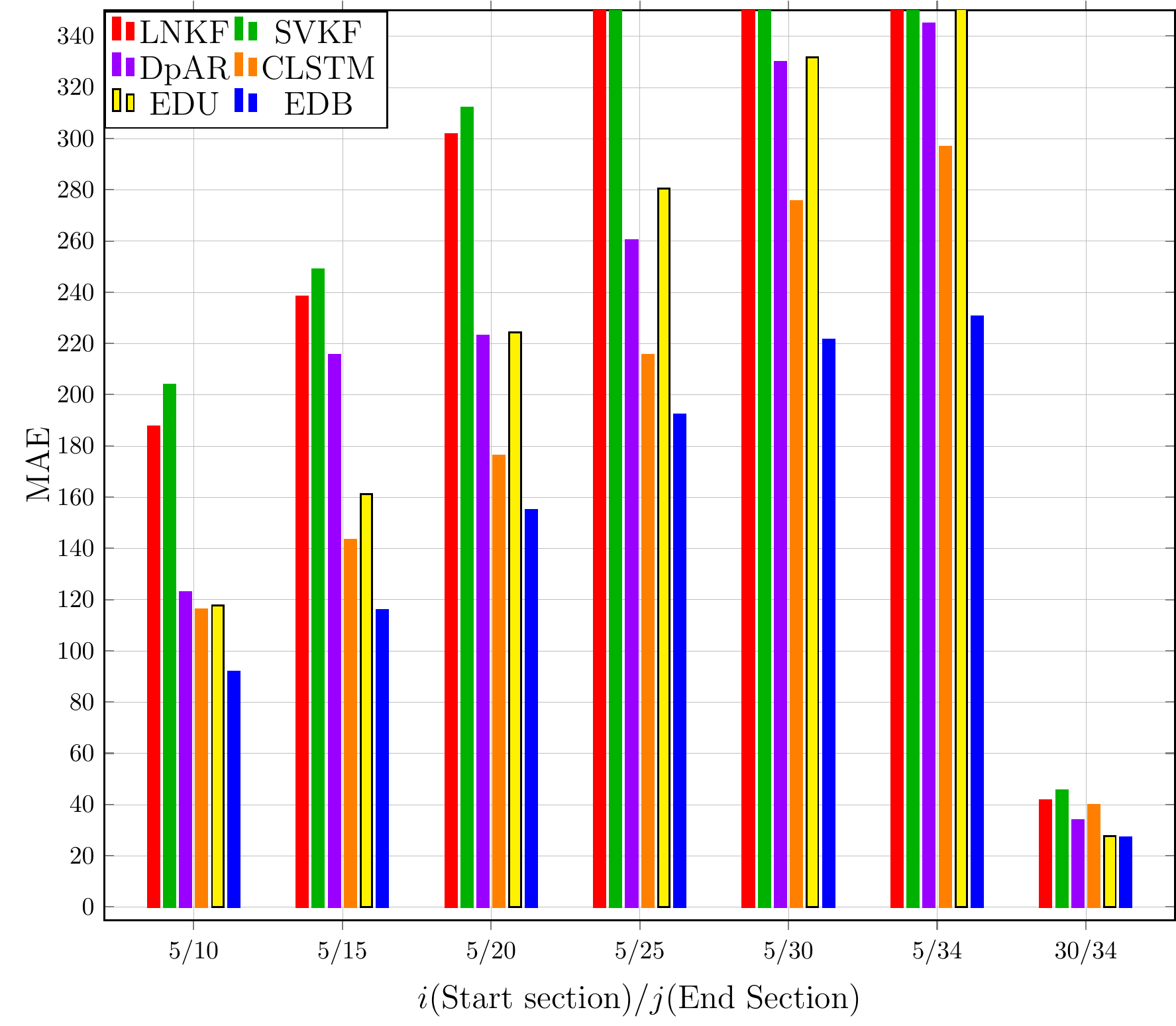}
	\label{fig:MAESec530} }
\subfigure[Bus current position at section $10$ and section $25$.]{\includegraphics[width=3.5in,height=2.6in]{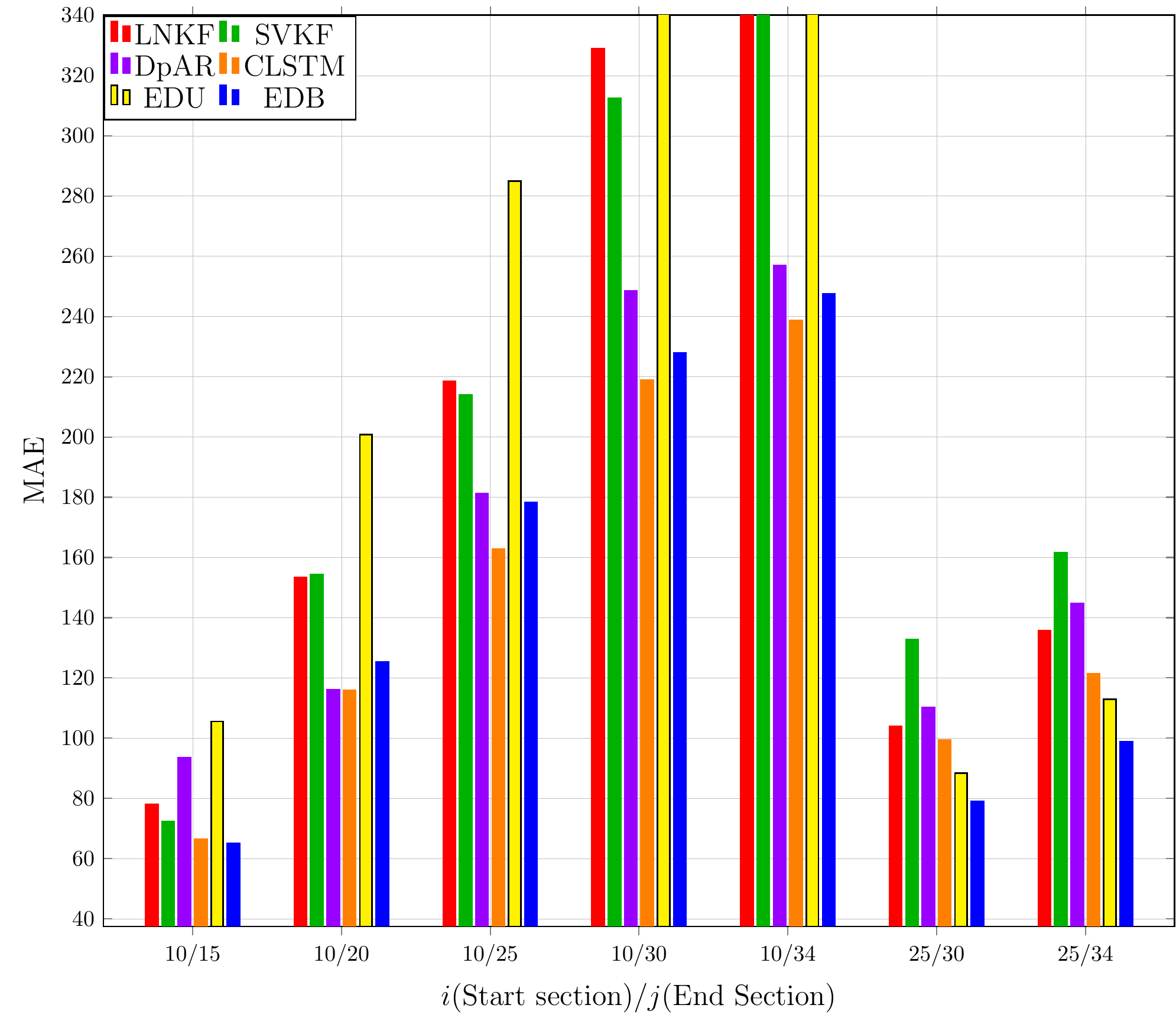}
	\label{fig:MAESec1025} }
\subfigure[Bus current position at section $15$ and section $20$.]{\includegraphics[width=3.5in,height=2.6in]{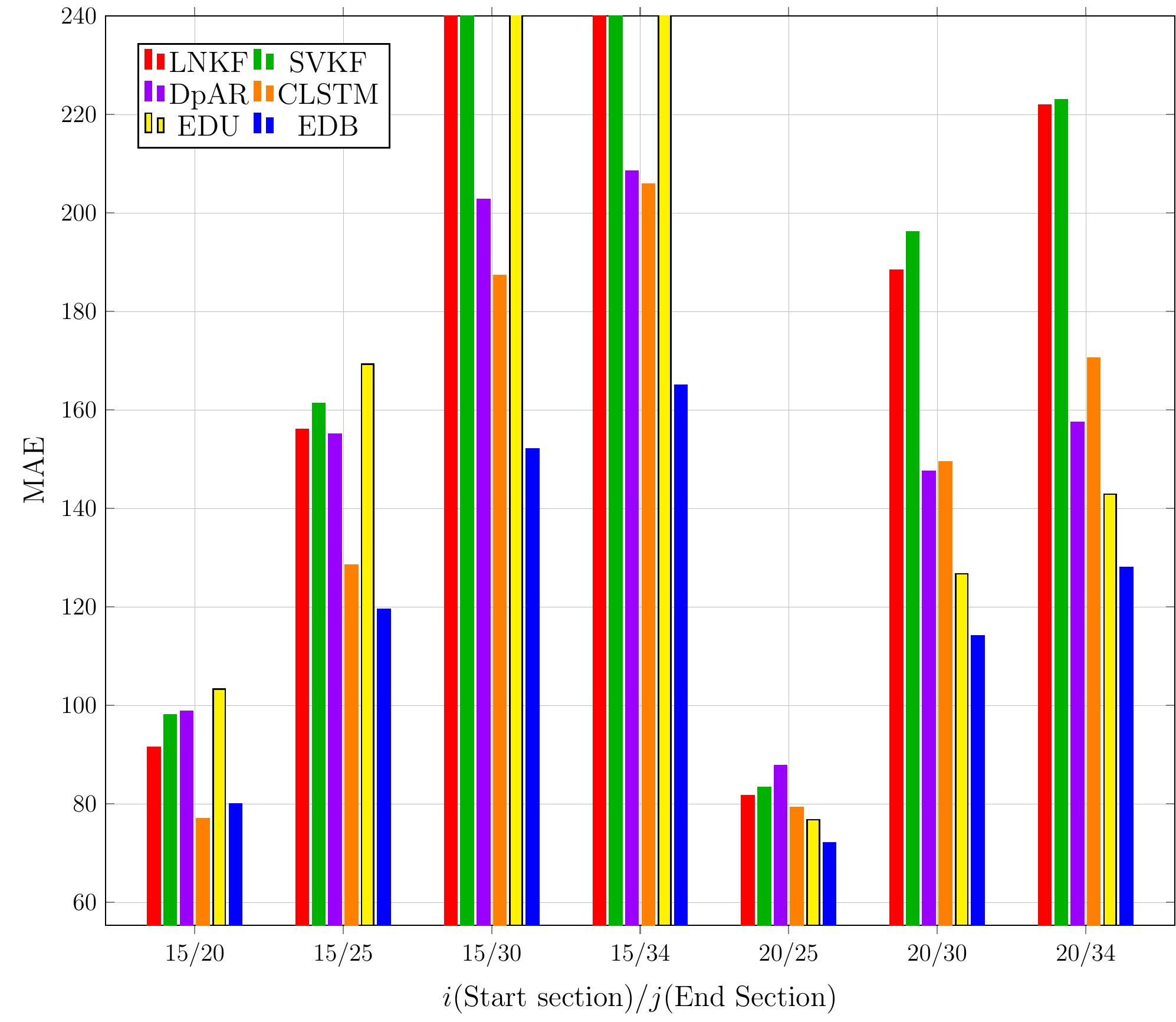}
	\label{fig:MAESec1520} }
	\caption{Multi-Step Ahead MAE at various start/current bus positions (again chosen in steps of 5) along the route.}
	\label{fig:MultistepMAE}
\end{figure}

We now test our learnt models on longer sub-routes, where sub-routes are spread across the entire route.  Our method must be able to provide quality  real-time predictions
 between any two bus-stops (or sections) of the bus route.   
We evaluate  performance based on both metrics.  Each sub-figure in Fig.~\ref{fig:MultistepMAPE}  shows comparison of MAPE values for an $(i,j)$ pair, where $i$ is the
current bus position and $j$ is the destination section.  This means for a given $i\in
\{5,10,15,20,25,30\}$,  $j$ is varied in steps of $5$ from $i+5$. The only exception here is  for the maximum $j$ which is $34$ (as the route ends there).
Fig.~\ref{fig:MAPESec530} provides MAPE results for $i=5,30$.  Fig.~\ref{fig:MAPESec1025} provides results for $i=10,25$, while Fig.~\ref{fig:MAPESec1520} provides results for 
$i=15,20$.
 MAE/MAPE is calculated by averaging across all trips and days in the test 
set by keeping the start section and end section fixed. 
We note that the proposed EDB  mostly performs better than   LNKF, SVKF, DpAR, CLSTM and EDU. 
This inference is   based on both MAPE and MAE metrics (Fig.~\ref{fig:MultistepMAPE} and Fig.~\ref{fig:MultistepMAE}).
EDB in particular, as in the
	2-step case performs the best. At the MAPE level, compared to CLSTM, of the 21 $(i,j)$ pairs considered, EDB outperforms CLSTM on $15$ $(i,j)$ pairs, has (statistically) similar  performance on $5$ $(i,j)$ pairs while does poorer than CLSTM on only one $(i,j)$ pair.  Compared to DpAR, EDB outperforms DpAR on $17$ $(i,j)$ pairs, has (statistically) similar  performance on $3$ $(i,j)$ pairs while does poorer than DpAR on only one $(i,j)$ pair. Compared to LNKF, SVKF and EDU, EDB outperforms all of them (statistically) on all $(i,j)$ pairs.  Specifically, for a $5$-step ahead prediction, EDB achieves MAPE improvement by 
up to 12.49\%, 12.92\%, 5.71\%, 3.71\%   and  8.61\% over LNKF, SVKF, DpAR, CLSTM  and EDU respectively.  For a $10$-step ahead prediction, EDB reduces MAPE by up 
to 9.83\%, 9.64\%, 8.45\%, 3.45\%  and 7.08\% in comparison  to LNKF, SVKF, DpAR,  CLSTM   and EDU respectively.
For a $15$-step ahead prediction, EDB reduces MAPE by up 
to 8.00\%,  7.93\%, 3.94\%, 2.61\%  and 6.59\%  in comparison  over LNKF, SVKF, DpAR, CLSTM  and EDU respectively.

On similar lines, Fig.~\ref{fig:MAESec530} provides MAE results for $i=5,30$.  Fig.~\ref{fig:MAESec1025} provides results for $i=10,25$, while Fig.~\ref{fig:MAESec1520} provides 
results for $i=15,20$. Please observe in these figures how the MAE generally increases with the length of the sub-route (i.e. $(j-i)$) inline with the intuition 
	that error increases with prediction horizon. As before, EDB performance is the best. The trends are similar as in the MAPE case. EDB does significantly better than CLSTM and DpAR on  $14$ and $18$ $(i,j)$ pairs respectively, while DpAR never does better and  CLSTM does better on only one $(i,j)$ pair. EDB as before outperforms LNKF, SVKF and EDU on all $(i,j)$ pairs. Specifically, for a $5$-step ahead prediction, EDB achieves MAE improvement by 
up to 95.89s, 112.04s, 31.15s, 24.3s   and  40.53s over LNKF, SVKF, DpAR, CLSTM    and EDU respectively.  For a $10$-step ahead prediction, EDB reduces MAE by up 
to 122.4s, 133.05s, 99.63s,  35.29s and 75.41s in comparison  over LNKF, SVKF, DpAR,  CLSTM     and EDU respectively.
For a $15$-step ahead prediction, EDB reduces MAE by up 
to 146.78s, 157.17s, 50.75s, 35.25s   and 106.76s in comparison  over LNKF, SVKF, DpAR, CLSTM  and EDU respectively.

Overall, our results indicate that our bidirectional variant, EDB, performs best on routes of varied lengths. Our results also indicate that EDU's performance is poor compared to some of the baselines and hence  not up-to the mark. This strongly vindicates the necessity  of a bidirectional layer at the decoder as in EDB, which 
potentially capture influence of past congestions from downstream sections on the route. 

\begin{remark}
In the plots whenever some of the baselines have high errors, we have avoided showing the associated bar heights completely as this adversely affects the visual 
	comparison of the bars in the error range of our proposed approaches EDB and EDU. As a result we see some clipped bars with actual error values higher than the 
	max $y$-axis range. 
\end{remark}

\section{Discussion \& Conclusion}
\label{sec:Conclusion}
In this paper we proposed a novel variant of Encoder Decoder (Seq2Seq) RNN approach with (i)bidirectional layer at the decoder and (ii)inputs carefully placed in a non-redundant fashion across the encoder and decoder, for Bus arrival time prediction (BATP). In particular,
\begin{itemize}
	\item To recognize that an ED based approach can be employed for real-time BATP by mapping
sequential aspect of ED to the spatial aspect of the problem, as proposed here, is not immediately evident.
{\em The geometry of real-time BATP, in particular the variable length feature of input-output training data enabled an interesting natural fit
with ED prediction framework to simultaneously predict travel times across subsequent
sections of the currently plying bus.}
\item We technically motivated in steps (in Sec.~\ref{sec:Prediction}  and Sec.~\ref{sec:EDArch} ) as
to how one can arrive at our novel ED architecture (with carefully chosen relevant  inputs)
incorporating both current spatio-temporal correlations and seasonal correlations. 
 The real-time spatio-temporal correlations and weekly 
seasonal influences were intelligently incorporated into the proposed predictive model via non-redundant inputs at both the encoder and decoder. 
	\item In the traditional ED, the only input at each step of the decoder is the same context vector
(which is the state of the last sequential-step of the encoder). While in our proposed ED,
we have additional distinct inputs at each step of the decoder coming from the closest (w.r.t
		the query time) previous bus and closest trip from previous week (same weekday). Since each step of
the decoder maps to a unique subsequent section, the (closest) previous bus’s entry time
and section travel time  are fed as a distinct input at the decoder step
associated with the same subsequent section (essentially in a space synchronized fashion).

	\item We propose a bidirectional layer at the decoder as this can now capture (for a given section) the possible influence of past congestions (in time) from the subsequent (downstream) sections
propagating backward in space (along the bus route). This novel feature of the bidirectional layer at the decoder is absent in both (i)traditional ED and (ii) other time- series
applications of the ED framework.	
		
\item We  demonstrated via detailed experiments  the utility of our approach  on a bus route from challenging mixed traffic
conditions by looking at prediction accuracy via two complementary metrics MAPE and MAE, along
sub-routes of a wide range of lengths. 
	We also ran tests of statistical significance to make sure
the improvements we are observing in favour of our method are indeed statistically significant.	
		Our EDB variant (which uses bidirectional layer at the
decoder) does better than the EDU variant (which uses unidirectional layer at the decoder). 
		and  many other carefully chosen state-of-art approaches for the same problem.  
\end{itemize}

{\bf Other Applications:} {\em The proposed ED framework 
with a bidirectional decoder can potentially have applications in multi-step time-series forecasting. For instance, it can be used  in multi-step sales prediction for domains like retail, CpG (Consume packaged goods) etc.} 
		Previous sales and price (that transpired in the recent past just before the forecast horizon) would be sequentially fed as encoder inputs. A $K$-step unfolded decoder's outputs match the sales targets of a $K$-step ahead forecast horizon. If future promotional information/prices across the multi-step forecast horizon is 
available, it could be fed as  decoder inputs in a synchronized fashion.  The point is that a buyer may decide to postpone buying certain products depending on the upcoming promotions and his current need, which means  
future promos can potentially influence past sales in the forecast horizon. This anti-causal influence can rightly be captured by our novel ED variant which uses a bidirectional variant at the decoder.  
In that sense, our proposed ED variant is
general and has other applications.  

As future work, we would explore utility of the
proposed architecture in domains like demand prediction for retail and so on. Another potential future work could be to exploit additional inputs for BATP like speed, flow
etc. and a suitably refined ED architecture to incorporate these inputs. We will also explore applying transformer and its variants to the encoder-decoder framework proposed in this paper for BATP.

\comment{
The proposed ED framework with a bidirectional decoder
can potentially have applications in sales prediction for retail
for instance. If promo information/prices is fed as inputs as
decoder inputs, then future promos can potentially influence
past sales. This anti-causal influence can rightly be captured
by our novel ED variant which uses a bidirectional variant
at the decoder. In that sense, our proposed framework is
general and can be employed in other time-series applications
as well. As future work, we would explore the utility of the
proposed architecture in other domains like demand prediction
for retail and so on.
}

\section*{Acknowledgement}
\noindent
        The authors thank Prof. Pravesh Biyani and his team from IIIT Delhi \cite{ITSC20} for kindly sharing real field AVL data, which made our bench-marking 
possible. This data is available on request for anyone. The authors also thank Rohith Regikumar (researcher, TCS Research) for his support during bench-marking.

\comment{
\section*{References}

\subsection*{Basic format for books:}\vspace*{-12pt}
\def\refname{}

\subsection*{Examples:}
\def\refname{}

\subsection*{Basic format for periodicals:}

\subsection*{Examples:}

\subsection*{Basic format for reports:}

\subsection*{Examples:}

\subsection*{Basic format for handbooks:}

\subsection*{Examples:}

\subsection*{Basic format for books (when available online):}

\subsection*{Examples:}

\subsection*{Basic format for journals (when available online):}

\subsection*{Examples:}

\subsection*{Basic format for papers presented at conferences (when available online):}

\subsection*{Example:}

\subsection*{Basic format for reports  and  handbooks (when available online):}

\subsection*{Examples:}

\subsection*{Basic format for computer programs and electronic documents (when available online):}

\subsection*{Example:}

\subsection*{Basic format for patents (when available online):}

\subsection*{Example:}

\subsection*{Basic format for conference proceedings (published):}

\subsection*{Example:}

\subsection*{Example for papers presented at conferences (unpublished):}

\subsection*{Basic format for patents:}

\subsection*{Example:}

\subsection*{Basic format for theses (M.S.) and dissertations (Ph.D.):}

\subsection*{Examples:}

\subsection*{Basic format for the most common types of unpublished references:}

\subsection*{Examples:}\vspace*{-6pt}

\subsection*{Basic formats for standards:}\vspace*{-6pt}

\subsection*{Examples:}\vspace*{-6pt}

\subsection*{Article number in reference examples:}\vspace*{-6pt}

\subsection*{Example when using et al.:}
\bibliographystyle{IEEEtai}
\bibliography{IEEEabrv,Transportation1}

\begin{thebibliography}{34}
\item[] J. K. Author, ``Title of chapter in the book,'' in {\em Title of His Published Book}, xth ed. City of Publisher, (only U.S. State), Country: Abbrev. of Publisher, year, ch. x, sec. x, pp. xxx--xxx.
\end{thebibliography}

\begin{thebibliography}{34}\vspace*{-12pt}

\bibitem{}G. O. Young, ``Synthetic structure of industrial plastics,'' in {\em Plastics},\break 2nd ed., vol. 3, J. Peters, Ed. New York, NY, USA: McGraw-Hill, 1964,\break pp. 15--64.

\bibitem{}W.-K. Chen, {\it Linear Networks and Systems}. Belmont, CA, USA: Wadsworth, 1993, pp. 123--135.

\end{thebibliography}

\vspace*{-12pt}

\begin{thebibliography}{34}
\item[]
J. K. Author, ``Name of paper,'' {\it Abbrev. Title of Periodical}, vol. {\it x},\break   no. {\it x}, pp. xxx--xxx, Abbrev. Month, year, DOI. \href{https://dx.doi.org/10.1109.XXX.123456}{10.1109.XXX.123456}.
\end{thebibliography}

\vspace*{-12pt}

\begin{thebibliography}{34}
\setcounter{enumiv}{2}

\bibitem{}J. U. Duncombe, ``Infrared navigation Part I: An assessment of feasibility,'' {\em IEEE Trans. Electron Devices}, vol. ED-11, no. 1, pp. 34--39,\break Jan. 1959, 10.1109/TED.2016.2628402.

\bibitem{}E. P. Wigner, ``Theory of traveling-wave optical laser,''
{\em Phys. Rev.},  vol.\break 134, pp. A635--A646, Dec. 1965. DOI. \href{https://dx.doi.org/10.1109.XXX.123456}{10.1109.XXX.123456}.

\bibitem{}E. H. Miller, ``A note on reflector arrays,'' {\em IEEE Trans. Antennas Propagat.}, to be published.
\end{thebibliography}

\vspace*{-12pt}
\begin{thebibliography}{34}
\item[]
J. K. Author, ``Title of report,'' Abbrev. Name of Co., City of Co., Abbrev. State, Country, Rep. xxx, year.
\end{thebibliography}

\vspace*{-12pt}
\begin{thebibliography}{34}
\setcounter{enumiv}{5}

\bibitem{} E. E. Reber, R. L. Michell, and C. J. Carter, ``Oxygen absorption in the earth’s atmosphere,'' Aerospace Corp., Los Angeles, CA, USA, Tech. Rep. TR-0200 (4230-46)-3, Nov. 1988.

\bibitem{} J. H. Davis and J. R. Cogdell, ``Calibration program for the 16-foot antenna,'' Elect. Eng. Res. Lab., Univ. Texas, Austin, TX, USA, Tech. Memo. NGL-006-69-3, Nov. 15, 1987.
\end{thebibliography}

\vspace*{-12pt}
\begin{thebibliography}{34}
\item[]
{\em Name of Manual/Handbook}, x ed., Abbrev. Name of Co., City of Co., Abbrev. State, Country, year, pp. xxx--xxx.
\end{thebibliography}

\vspace*{-12pt}

\begin{thebibliography}{34}
\setcounter{enumiv}{7}

\bibitem{} {\em Transmission Systems for Communications}, 3rd ed., Western Electric Co., Winston-Salem, NC, USA, 1985, pp. 44--60.

\bibitem{} {\em Motorola Semiconductor Data Manual}, Motorola Semiconductor Products Inc., Phoenix, AZ, USA, 1989.
\end{thebibliography}

\vspace*{-12pt}
\begin{thebibliography}{34}
\item[]
J. K. Author, ``Title of chapter in the book,'' in {\em Title of Published Book}, xth ed. City of Publisher, State, Country: Abbrev. of Publisher, year, ch. x, sec. x, pp. xxx--xxx. [Online]. Available: http://www.web.com 
\end{thebibliography}

\vspace*{-12pt}

\begin{thebibliography}{34}
\setcounter{enumiv}{9}

\bibitem{}G. O. Young, ``Synthetic structure of industrial plastics,'' in Plastics, vol. 3, Polymers of Hexadromicon, J. Peters, Ed., 2nd ed. New York, NY, USA: McGraw-Hill, 1964, pp. 15--64. [Online]. Available: http://www.bookref.com. 

\bibitem{} {\em The Founders Constitution}, Philip B. Kurland and Ralph Lerner, eds., Chicago, IL, USA: Univ. Chicago Press, 1987. [Online]. Available: http://press-pubs.uchicago.edu/founders/

\bibitem{} The Terahertz Wave eBook. ZOmega Terahertz Corp., 2014. [Online]. Available: http://dl.z-thz.com/eBook/zomega\_ebook\_pdf\_1206\_sr.pdf. Accessed on: May 19, 2014. 

\bibitem{} Philip B. Kurland and Ralph Lerner, eds., {\em The Founders Constitution}. Chicago, IL, USA: Univ. of Chicago Press, 1987, Accessed on: Feb. 28, 2010, [Online] Available: http://press-pubs.uchicago.edu/founders/ 
\end{thebibliography}

\vspace*{-12pt}
\begin{thebibliography}{34}
\item[] J. K. Author, ``Name of paper,'' {\em Abbrev. Title of Periodical}, vol. x, no. x, pp. xxx--xxx, Abbrev. Month, year. Accessed on: Month, Day, year, DOI: 10.1109.XXX.123456, [Online].
\end{thebibliography}

\vspace*{-12pt}

\begin{thebibliography}{34}
\setcounter{enumiv}{13}

\bibitem{}J. S. Turner, ``New directions in communications,'' {\em IEEE J. Sel. Areas Commun.}, vol. 13, no. 1, pp. 11--23, Jan. 1995. 

\bibitem{} W. P. Risk, G. S. Kino, and H. J. Shaw, ``Fiber-optic frequency shifter using a surface acoustic wave incident at an oblique angle,'' {\em Opt. Lett.}, vol. 11, no. 2, pp. 115--117, Feb. 1986.

\bibitem{} P. Kopyt {\em et al.}, ``Electric properties of graphene-based conductive layers from DC up to terahertz range,'' {\em IEEE THz Sci. Technol.}, to be published. DOI: \href{https://dx.doi.org/10.1109.XXX.123456}{10.1109/TTHZ.2016.2544142}.
\end{thebibliography}

\vspace*{-12pt}
\begin{thebibliography}{34}
\item[] J.K. Author. (year, month). Title. presented at abbrev. conference title. [Type of Medium]. Available: site/path/file
\end{thebibliography}

\vspace*{-12pt}

\begin{thebibliography}{34}
\setcounter{enumiv}{16}

\bibitem{}PROCESS Corporation, Boston, MA, USA. Intranets: Internet technologies deployed behind the firewall for corporate productivity. Presented at INET96 Annual Meeting. [Online]. Available: http://home.process.com/Intranets/wp2.htp
\end{thebibliography}

\vspace*{-12pt}
\begin{thebibliography}{34}
\item[] J. K. Author. ``Title of report,'' Company. City, State, Country. Rep. no., (optional: vol./issue), Date. [Online] Available: site/path/file 
\end{thebibliography}

\vspace*{-12pt}

\begin{thebibliography}{34}
\setcounter{enumiv}{17}

\bibitem{}R. J. Hijmans and J. van Etten, ``Raster: Geographic analysis and modeling with raster data,'' R Package Version 2.0-12, Jan. 12, 2012. [Online]. Available: http://CRAN.R-project.org/package=raster 

\bibitem{}Teralyzer. Lytera UG, Kirchhain, Germany [Online]. Available: http://www.lytera.de/Terahertz\_THz\_Spectroscopy.php?id=home, Accessed on: Jun. 5, 2014.
\end{thebibliography}

\vspace*{-12pt}
\begin{thebibliography}{34}
\item[] Legislative body. Number of Congress, Session. (year, month day). {\em Number of bill or resolution, Title}. [Type of medium]. Available: site/path/file

\item[] {\em NOTE:} ISO recommends that capitalization follow the accepted practice for the language or script in which the information is given.
\end{thebibliography}

\vspace*{-12pt}

\begin{thebibliography}{34}
\setcounter{enumiv}{19}

\bibitem{}U. S. House. 102nd Congress, 1st Session. (1991, Jan. 11). {\em H. Con. Res. 1, Sense of the Congress on Approval of Military Action}. [Online]. Available: LEXIS Library: GENFED File: BILLS 
\end{thebibliography}

\vspace*{-12pt}
\begin{thebibliography}{34}
\item[] Name of the invention, by inventor’s name. (year, month day). Patent Number [Type of medium]. Available:site/path/file
\end{thebibliography}

\vspace*{-12pt}

\begin{thebibliography}{34}
\setcounter{enumiv}{20}

\bibitem{}Musical tooth brush with mirror, by L. M. R. Brooks. (1992, May 19). Patent D 326 189
[Online]. Available: NEXIS Library: LEXPAT File:   DES 
\end{thebibliography}

\vspace*{-12pt}
\begin{thebibliography}{34}
\item[] J. K. Author, ``Title of paper,'' in {\em Abbreviated Name of Conf.}, City of Conf., Abbrev. State (if given), Country, year, pp. xxx--xxx.
\end{thebibliography}

\vspace*{-12pt}

\begin{thebibliography}{34}
\setcounter{enumiv}{21}

\bibitem{}D. B. Payne and J. R. Stern, ``Wavelength-switched passively coupled single-mode optical network,'' in {\em Proc. IOOC-ECOC}, Boston, MA, USA, 1985,
pp. 585--590. 

\end{thebibliography}

\vspace*{-12pt}

\begin{thebibliography}{34}
\setcounter{enumiv}{22}

\bibitem{}D. E behard and E. Voges, ``Digital single sideband detection for inter ferometric sensors,'' presented at the {\em 2nd Int. Conf. Optical Fiber Sensors}, Stuttgart, Germany, Jan. 2--5, 1984.
\end{thebibliography}

\vspace*{-12pt}
\begin{thebibliography}{34}
\item[] J. K. Author, ``Title of patent,'' U.S. Patent x xxx--xxx, Abbrev. Month, day, year.
\end{thebibliography}

\vspace*{-12pt}

\begin{thebibliography}{34}
\setcounter{enumiv}{23}

\bibitem{}G. Brandli and M. Dick, ``Alternating current fed power supply,'' U.S. Patent 4 084 217, Nov. 4, 1978.
\end{thebibliography}

\vspace*{-12pt}
\begin{thebibliography}{34}
\item[a)] J. K. Author, ``Title of thesis,'' M.S. thesis, Abbrev. Dept., Abbrev. Univ., City of Univ., Abbrev. State, year.

\item[b)] J. K. Author, ``Title of dissertation,'' Ph.D. dissertation, Abbrev. Dept., Abbrev. Univ., City of Univ., Abbrev. State, year.
\end{thebibliography}

\vspace*{-12pt}

\begin{thebibliography}{34}
\setcounter{enumiv}{24}

\bibitem{}J. O. Williams, ``Narrow-band analyzer,'' Ph.D. dissertation, Dept. Elect. Eng., Harvard Univ., Cambridge, MA, USA, 1993.

\bibitem{}N. Kawasaki, ``Parametric study of thermal and chemical nonequilibrium nozzle flow,'' M.S. thesis, Dept. Electron. Eng., Osaka Univ., Osaka, Japan, 1993.
\end{thebibliography}

\vspace*{-12pt}
\begin{thebibliography}{34}
\item[a)] J. K. Author, private communication, Abbrev. Month, year.

\item[b)] J. K. Author, ``Title of paper,'' unpublished.

\item[c)] J. K. Author, ``Title of paper,'' to be published.
\end{thebibliography}

\vspace*{-18pt}

\begin{thebibliography}{34}
\setcounter{enumiv}{26}

\bibitem{}A. Harrison, private communication, May 1995.

\bibitem{}B. Smith, ``An approach to graphs of linear forms,'' unpublished.

\bibitem{}A. Brahms, ``Representation error for real numbers in binary computer arithmetic,'' IEEE Computer Group Repository, Paper R-67-85.
\end{thebibliography}

\vspace*{-18pt}
\begin{thebibliography}{34}
\item[a)] {\em Title of Standard}, Standard number, date.

\item[b)] {\em Title of Standard}, Standard number, Corporate author, location, date.
\end{thebibliography}

\vspace*{-18pt}

\begin{thebibliography}{34}
\setcounter{enumiv}{29}
\bibitem{}IEEE Criteria for Class IE Electric Systems, IEEE Standard 308, 1969.

\bibitem{} Letter Symbols for Quantities, ANSI Standard Y10.5-1968.
\end{thebibliography}

\vspace*{-18pt}

\begin{thebibliography}{34}

\setcounter{enumiv}{31}

\bibitem{}R. Fardel, M. Nagel, F. Nuesch, T. Lippert, and A. Wokaun, ``Fabrication of organic light emitting diode pixels by laser-assisted forward transfer,'' {\em Appl. Phys. Lett.}, vol. 91, no. 6, Aug. 2007, Art. no. 061103. 

\bibitem{} J. Zhang and N. Tansu, ``Optical gain and laser characteristics of InGaN quantum wells on ternary InGaN substrates,'' {\em IEEE Photon. J.}, vol. 5, no. 2, Apr. 2013, Art. no. 2600111.  \href{https://dx.doi.org/10.1109.XXX.123456}{10.1109.XXX.123456}.
\end{thebibliography}

\vspace*{-18pt}

\begin{thebibliography}{34}
\setcounter{enumiv}{33}

\bibitem{}S. Azodolmolky {\em et al.}, ``Experimental demonstration of an impairment aware network planning and operation tool for transparent/translucent optical networks,'' {\em J. Lightw. Technol.}, vol. 29, no. 4, pp. 439--448, Sep. 2011.
\end{thebibliography}

\begin{thebibliography}{34}




\bibitem{GoogleETA21}
A.~Derrow{-}Pinion, J.~She, D.~Wong, O.~Lange, T.~Hester, L.~Perez,
  M.~Nunkesser, S.~Lee, X.~Guo, B.~Wiltshire, P.~W. Battaglia, V.~Gupta, A.~Li,
  Z.~Xu, A.~Sanchez{-}Gonzalez, Y.~Li, and P.~Velickovic, ``{ETA} prediction
  with graph neural networks in google maps,'' in \emph{{CIKM}}.\hskip 1em plus
  0.5em minus 0.4em\relax {ACM}, 2021, pp. 3767--3776.

\bibitem{KDD20}
R.~Barnes, S.~Buthpitiya, J.~Cook, A.~Fabrikant, A.~Tomkins, and F.~Xu,
  ``Bustr: Predicting bus travel times from real-time traffic,'' in \emph{KDD
  '20: Proceedings of the 26th ACM SIGKDD International Conference on Knowledge
  Discovery Data Mining}.\hskip 1em plus 0.5em minus 0.4em\relax {ACM}, 2020,
  pp. 3243--3251.

\bibitem{ranji19}
P.~Ranjitkar, L.~S. Tey, E.~Chakravorty, and K.~L. Hurley, ``Bus arrival time
  modeling based on auckland data,'' \emph{Transportation Research Record},
  vol. 2673, no.~6, pp. 1--9, 2019.

\bibitem{Achar19a}
A.~{Achar}, D.~{Bharathi}, B.~A. {Kumar}, and L.~{Vanajakshi}, ``Bus arrival
  time prediction: A spatial kalman filter approach,'' \emph{IEEE Transactions
  on Intelligent Transportation Systems}, vol.~21, no.~3, pp. 1298--1307, 2020.

\bibitem{Paliwal19}
C.~{Paliwal} and P.~{Biyani}, ``To each route its own eta: A generative
  modeling framework for eta prediction,'' in \emph{2019 IEEE Intelligent
  Transportation Systems Conference (ITSC)}, Oct 2019, pp. 3076--3081.

\bibitem{lelitha:09}
L.~Vanajakshi, S.~C. Subramanian, and R.~Sivanandan, ``Travel time prediction
  under heterogeneous traffic conditions using global positioning system data
  from buses,'' \emph{IET Intelligent Transporation Systems}, vol. 3(1), pp.
  1--9, 2009.

\bibitem{Jairam18}
R.~Jairam, B.~A. Kumar, S.~S. Arkatkar, and L.~Vanajakshi, ``Performance
  comparison of bus travel time prediction models across indian cities,''
  \emph{Transportation Research Record}, vol. 2672, no.~31, pp. 87--98, 2018.

\bibitem{Dhivya20}
B.~D. Bharathi, B.~A. Kumar, A.~Achar, and L.~Vanajakshi, ``Bus travel time
  prediction: a log-normal auto-regressive (ar) modelling approach,''
  \emph{Transportmetrica A: Transport Science}, vol.~16, no.~3, pp. 807--839,
  2020.

\bibitem{bin:07}
Y.~Bin, Y.~Zhinzhen, and Y.~Baozhen, ``Bus arrival time prediction using
  support vector machines,'' \emph{Journal of Intelligent Transportation
  Systems}, vol. 10(4), pp. 151--158, 2007.

\bibitem{reddy:16}
K.~K. Reddy, B.~A. Kumar, and L.~Vanajakshi, ``Bus travel time prediction under
  high variability conditions,'' \emph{Current Science}, vol. 111, no.~4, pp.
  700--711, 2016.

\bibitem{Achar19b}
A.~{Achar}, R.~{Regikumar}, and B.~A. {Kumar}, ``Dynamic bus arrival time
  prediction exploiting non-linear correlations,'' in \emph{2019 International
  Joint Conference on Neural Networks (IJCNN)}, July 2019, pp. 1--8.

\bibitem{fan:15}
W.~Fan and Z.~Gurmu, ``Dynamic travel time prediction models for buses using
  only gps data,'' \emph{International Journal of Transportation Science and
  Technology}, vol.~4, no.~4, pp. 353 -- 366, 2015.

\bibitem{Petersen19}
N.~C. Petersen, F.~Rodrigues, and F.~C. Pereira, ``Multi-output bus travel time
  prediction with convolutional lstm neural network,'' \emph{Expert Syst.
  Appl.}, vol. 120, pp. 426--435, 2019.

\bibitem{Ran19}
X.~Ran, Z.~Shan, Y.~Fang, and C.~Lin, ``An lstm-based method with attention
  mechanism for travel time prediction,'' \emph{Sensors}, vol.~19, no.~4, 2019.
  [Online]. Available: \url{https://www.mdpi.com/1424-8220/19/4/861}

\bibitem{Vignesh20}
L.~K.~P. Vignesh, A.~Achar, and G.~Karthik, ``Dynamic bus arrival time
  prediction: A temporal difference learning approach,'' in \emph{2020
  International Joint Conference on Neural Networks (IJCNN)}, 2020, pp. 1--8.

\bibitem{snig:15}
B.~A. Kumar, S.~Mothukuri, L.~Vanajakshi, and S.~C. Subramanian, ``Analytical
  approach to identify the optimum inputs for a bus travel time prediction
  method,'' \emph{Transportation Research Record: Journal of the Transportation
  Research Board}, vol. 2535, pp. 25--34, 2015.

\bibitem{partc:17}
B.~A. Kumar, L.~Vanajakshi, and S.~C. Subramanian, ``Bus travel time prediction
  using a time-space discretization approach,'' \emph{Transportation Research
  Part C: Emerging Technologies}, vol.~79, pp. 308--332, 2017.

\bibitem{vivek:17}
B.~A. Kumar, V.~Kumar, L.~Vanajakshi, and S.~Subramanian, ``Performance
  comparison of data driven and less data demanding techniques for bus travel
  time prediction,'' \emph{European Transport}, vol. 65(9), 2017.

\bibitem{Duan16}
Y.~{Duan}, Y.~{L.V.}, and F.~{Wang}, ``Travel time prediction with lstm neural
  network,'' in \emph{2016 IEEE 19th International Conference on Intelligent
  Transportation Systems (ITSC)}, Nov 2016, pp. 1053--1058.

\bibitem{yang:16}
M.~Yang, C.~C. ., L.~Wangz., X.~Yanx., and L.~Zhou, ``Bus arrival time
  prediction using support vector machine with genetic algorithm,''
  \emph{Neural Network World}, vol.~3, pp. 205--217, 2016.

\bibitem{Wen17}
R.~Wen, K.~Torkkola, B.~Narayanaswamy, and D.~Madeka, ``A multi-horizon
  quantile recurrent forecaster,'' in \emph{The 31st Conference on Neural
  Information Processing Systems (NIPS 2017), Time Series Workshop}, 2017.

\bibitem{Yagmur17}
Y.~G. Cinar, H.~Mirisaee, P.~Goswami, {\'{E}}.~Gaussier,
  A.~A{\"{\i}}t{-}Bachir, and V.~V. Strijov, ``Position-based content attention
  for time series forecasting with sequence-to-sequence rnns,'' in \emph{Neural
  Information Processing - 24th International Conference, {ICONIP} 2017,
  Guangzhou, China, November 14-18, 2017, Proceedings, Part {V}}, ser. Lecture
  Notes in Computer Science, vol. 10638.\hskip 1em plus 0.5em minus 0.4em\relax
  Springer, 2017, pp. 533--544. 

\bibitem{Cho14}
K.~Cho, B.~van Merri{\"e}nboer, C.~Gulcehre, D.~Bahdanau, F.~Bougares,
  H.~Schwenk, and Y.~Bengio, ``Learning phrase representations using {RNN}
  encoder{--}decoder for statistical machine translation,'' in
  \emph{Proceedings of the 2014 Conference on Empirical Methods in Natural
  Language Processing ({EMNLP})}, Oct. 2014, pp. 1724--1734.

\bibitem{Sutskever14}
I.~Sutskever, O.~Vinyals, and Q.~V. Le, ``Sequence to sequence learning with
  neural networks,'' in \emph{Proceedings of the 27th International Conference
  on Neural Information Processing Systems - Volume 2}, 2014, p. 3104–3112.

\bibitem{lin:99}
W.~H. Lin and J.~Zeng, ``Experimental study on real-time bus arrival time
  prediction with {GPS} data,'' \emph{Transportation Research Record: Journal
  of the Transportation Research Board}, vol. 1666, pp. 101--109, 1999.

\bibitem{zhou:12}
P.~Zhou, Y.~Zheng, and M.~Li, ``How long to wait?: Predicting bus arrival time
  with mobile phone based participatory sensing,'' in \emph{Proceedings of the
  10th International Conference on Mobile Systems, Applications, and
  Services}.\hskip 1em plus 0.5em minus 0.4em\relax New York, NY, USA: ACM,
  2012, pp. 379--392.

\bibitem{shalaby:03}
A.~Shalaby and A.~Farhan, ``Bus travel time prediction for dynamic operations
  control and passenger information systems,'' in \emph{82nd Annual Meeting of
  the Transportation Research Board}.\hskip 1em plus 0.5em minus 0.4em\relax
  Washington D.C., USA: National Research Council, 2003.

\bibitem{kumar:12}
S.~V. Kumar and L.~Vanajakshi, ``Pattern identification based bus arrival time
  prediction,'' \emph{Proceedings of the Institution of Civil
  Engineers-Transport}, vol. 167(3), pp. 194--203, 2012.

\bibitem{sun:07}
D.~Sun, H.~Luo, L.~Fu, W.~Liu, X.~Liao, and M.~Zhao, ``Predicting bus arrival
  time on the basis of global positioning system data,'' \emph{Transportation
  Research Record: Journal of the Transportation Research Board}, vol. 2034,
  pp. 62--72, 2007.

\bibitem{Wu20}
J.~Wu, Q.~Wu, J.~Shen, and C.~Cai, ``Towards attention-based convolutional long
  short-term memory for travel time prediction of bus journeys,''
  \emph{Sensors}, vol.~20, no.~12, 2020. 

\bibitem{Liang15}
L.~Lu, X.~Zhang, K.~Cho, and S.~Renals, {A
  study of the recurrent neural network encoder-decoder for large vocabulary
  speech recognition},'' in \emph{Proceedings
  of the 16th Annual Conference of the International Speech Communication
  Association, INTERSPEECH}, Jan. 2015, pp. 3249--3253.

\bibitem{kumar:13}
B.~A. Kumar, L.~Vanjakshi, and S.~C. Subramanian, ``Day-wise travel time
  pattern analysis under heterogeneous traffic conditions,'' \emph{Procedia -
  Social and Behavioral Sciences}, vol. 104, pp. 746 -- 754, 2013, 2nd
  Conference of Transportation Research Group of India (2nd CTRG).

\bibitem{Pennington14}
J.~Pennington, R.~Socher, and C.~Manning, ``{G}lo{V}e: Global vectors for word
  representation,'' in \emph{Proceedings of the 2014 Conference on Empirical
  Methods in Natural Language Processing ({EMNLP})}.\hskip 1em plus 0.5em minus
  0.4em\relax Doha, Qatar: Association for Computational Linguistics, Oct.
  2014, pp. 1532--1543. 

\bibitem{Mikolov13}
T.~Mikolov, K.~Chen, G.~Corrado, and J.~Dean, ``Efficient estimation of word
  representations in vector space,'' \emph{CoRR}, vol. abs/1301.3781, 2013.

\bibitem{Hochreiter97}
S.~Hochreiter and J.~Schmidhuber, ``Long short-term memory,'' \emph{Neural
  Comput.}, vol.~9, no.~8, p. 1735–1780, Nov. 1997.

\bibitem{Chung14}
J.~Chung, C.~Gulcehre, K.~Cho, and Y.~Bengio, ``Empirical evaluation of gated
  recurrent neural networks on sequence modeling,'' in \emph{NIPS 2014 Deep
  Learning and Representation Learning Workshop}, 2014. 

\bibitem{Gupta17}
A.~Gupta, G.~Gurrala, and P.~S. Sastry, ``Instability prediction in power
  systems using recurrent neural networks,'' in \emph{Proceedings of the 26th
  International Joint Conference on Artificial Intelligence}.\hskip 1em plus
  0.5em minus 0.4em\relax AAAI Press, 2017, p. 1795–1801.

\bibitem{Ravanelli18}
M.~Ravanelli, P.~Brakel, M.~Omologo, and Y.~Bengio, ``Light gated recurrent
  units for speech recognition,'' \emph{IEEE Transactions on Emerging Topics in
  Computational Intelligence}, vol.~2, no.~2, p. 92–102, Apr 2018.

\bibitem{Che16}
Z.~Che, S.~Purushotham, K.~Cho, D.~Sontag, and Y.~Liu, ``Recurrent neural
  networks for multivariate time series with missing values,'' \emph{Scientific
  Reports}, vol.~8, Jun 2016.

\bibitem{Nicole20}
N.~Gruber and A.~Jockisch, ``Are gru cells more specific and lstm cells more
  sensitive in motive classification of text?'' \emph{Front. Artif. Intell.},
  2020.

\bibitem{Flunkert17}
D.~Salinas, V.~Flunkert, J.~Gasthaus, and T.~Januschowski, ``Deepar:
  Probabilistic forecasting with autoregressive recurrent networks,''
  \emph{International Journal of Forecasting}, vol.~36, no.~3, pp. 1181--1191,
  2020.

\bibitem{ITSC20}
A.~Kumar, S.~Balodi, A.~Jain, and P.~Biyani, ``Benchmark dataset for timetable
  optimization of bus routes in the city of new delhi,'' in \emph{IEEE 23rd
  International Conference on Intelligent Transportation Systems (ITSC)}.\hskip
  1em plus 0.5em minus 0.4em\relax {IEEE}, 2020, pp. 1--6.

\bibitem{mihaylova:04}
L.Mihaylova and R.Boel, ``A particle filter for freeway traffic estimation,''
  in \emph{43rd IEEE Conference on Decision and Control}, 2004.

\bibitem{mori:15}
U.~Mori, A.~Mendiburu, M.~Alvarez, and J.~Lozano, ``A review of travel time
  estimation and forecasting for advanced traveller information systems,''
  \emph{Transportmetrica A: Transport Science}, vol.~11, no.~2, pp. 119--157,
  2015.

\bibitem{zhang:09}
J.~Zhang, L.~Yan, Y.~Han, and J.~Zhang, ``Study on the prediction model of bus
  arrival time,'' in \emph{International Conference on Management and Service
  Science}, MASS, China, 2009, pp. 1--3.

\bibitem{xinghao:13}
S.~Xinghao, T.~Jing, C.~Guojun, and S.~Qichong, ``Predicting bus real-time
  travel time basing on both {GPS} and {RFID} data,'' \emph{Procedia - Social
  and Behavioral Sciences}, vol.~96, pp. 2287--2299, 2013.



\end{thebibliography}
}



\appendices

\comment{
\section{Traffic-theory based methods}
\label{app:TT}
These methods try to explicitly model the dynamics/physics of the traffic.
The associated models could broadly be categorized either as (a)macroscopic (modeling at a coarse level capturing aggregate variables flow,
density and speed) \cite{mihaylova:04} OR (b)microscopic (detailed modeling at a vehicular level) \cite{mori:15}. While the macroscopic models give only an aggregate information,
microscopic
models suffer from issues like  being computationally expensive, necessity to  calibrate,
 and making inaccurate  predictions.
A recent approach based on a traffic-theory based
model using speed as input was proposed in \cite{partc:17}. \cite{zhang:09} breaks the travel time into different components like queuing delay at signals, acceleration and
deceleration into and out of bus-stops which incorporates the physics of traffic into prediction. \cite{xinghao:13} does a speed-based prediction and finally uses some traffic
theory based ideas to convert speed into travel times.
}

\section{Route Visualization on Map}
\label{app:Route}
\begin{figure*}[!h]
\center
  \includegraphics[height=4.0in, width=4.5in]{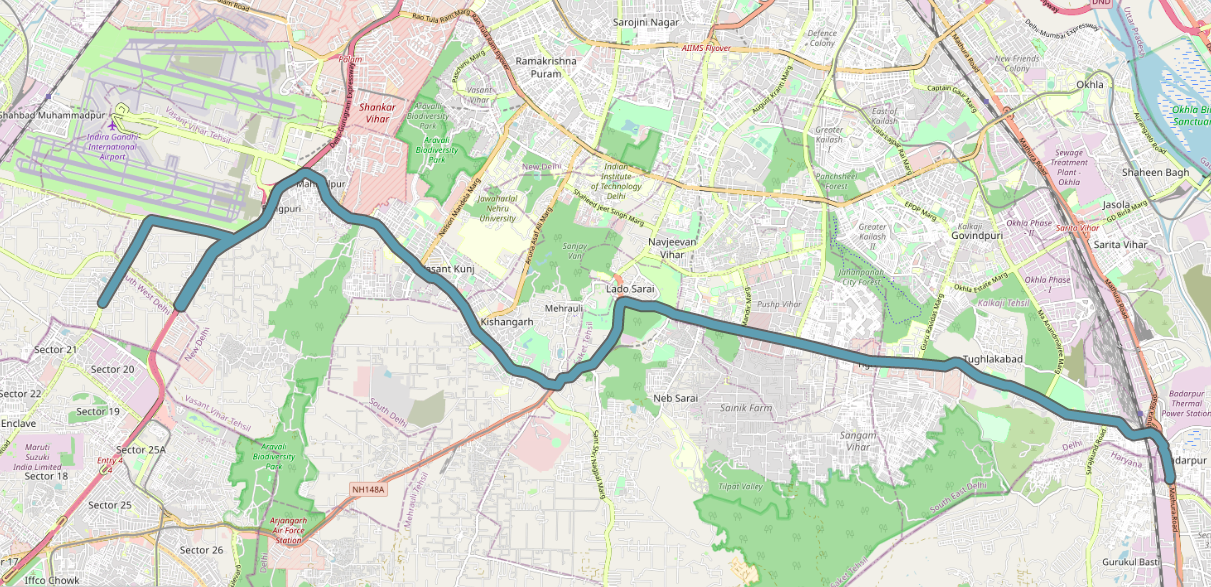}
\caption{Route 399}
\label{fig:Route399}
\vspace{-0.00in}
\end{figure*}

\comment{
\begin{IEEEbiography}[{\includegraphics[width=1.4in,height=1.3in,clip,keepaspectratio]{Nancy}}]
	{Nancy Bhutani}{\space}
received her B.Sc. Hons. degree in Mathematics from Delhi University.  She has a masters in Mathematics from Indian Institute of Technology Bhubaneswar.
Currently,  she is a researcher at TCS research, Chennai. Her research interests mainly include machine learning and deep learning.

\end{IEEEbiography}

\begin{IEEEbiography}[{\includegraphics[width=1.4in,height=1.3in,clip,keepaspectratio]{Soumen}}]
{Soumen Pachal} received his M.Tech degree in computer science from Indian Statistical Institute, India. He also has an M.Sc degree in mathematics from
University of Calcutta. Currently he is working at TCS Research, India as a researcher. His research interests are  mainly in machine learning and deep learning.
   
\end{IEEEbiography}

\begin{IEEEbiography}[{\includegraphics[width=1.4in,height=1.3in,clip,keepaspectratio]{Avinash}}]
{Avinash Achar}
        got his B.Tech from National Institute of Technology, Karnataka. He has a masters and Ph.D from the Indian Institute of Science, Bangalore. He was an ERCIM post-doctoral fellow at NTNU, Norway. He is currently a Senior Research Scientist at TCS Research, Chennai. His research interests broadly span the areas of data mining, machine learning and their applications in different domains.

\end{IEEEbiography}
}

\end{document}